\documentclass{article}

\usepackage{arxiv}

\usepackage[utf8]{inputenc} 
\usepackage[T1]{fontenc}    
\usepackage{hyperref}       
\usepackage{url}            
\usepackage{booktabs}       
\usepackage{amsfonts}       
\usepackage{nicefrac}       
\usepackage{microtype}      
\usepackage{lipsum}		
\usepackage{graphicx}
\usepackage[authoryear,semicolon]{natbib}
\usepackage{doi}

\usepackage{amsmath}    
\usepackage{amssymb}    
\DeclareMathOperator{\atantwo}{atan2}
\DeclareMathOperator*{\E}{\mathbb{E}}  
\DeclareMathOperator*{\modd}{\scriptstyle\%}

\usepackage{algorithmic}

\usepackage{array}

\title{Unscented Kalman Filter for Long-Distance Vessel Tracking in Geodetic Coordinates}

\date{November 25, 2021}	

\author{ \href{https://orcid.org/0000-0002-4957-6903}{\includegraphics[scale=0.06]{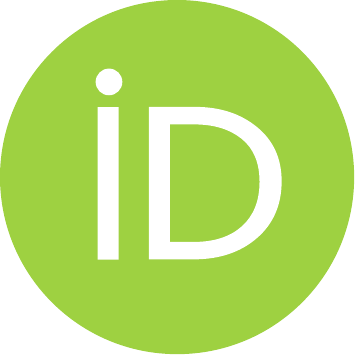}\hspace{1mm}Blake Cole}\thanks{B. Cole is with the Laboratory for Autonomous Marine Sensing Systems, Department of Mechanical Engineering.} \\
	Department of Mechanical Engineering\\
	Massachusetts Institute of Technology\\
	Cambridge, MA 02139 \\
	\texttt{blerk@mit.edu} \\
	\And
	\href{https://orcid.org/0000-0002-4188-9614}{\includegraphics[scale=0.06]{orcid.pdf}\hspace{1mm}Gabriel Schamberg}\thanks{G. Schamberg is with the Neuroscience Statistics Research Laboratory, Picower Institute for Learning and Memory.} \\
	Picower Institute for Learning and Memory\\
	Massachusetts Institute of Technology\\
	Cambridge, MA 02139 \\
	\texttt{gabes@mit.edu} \\
}

\renewcommand{\shorttitle}{Geodetic Unscented Kalman Filter}

\hypersetup{
pdftitle={Geodetic Unscented Kalman Filter},
pdfsubject={q-bio.NC, q-bio.QM},
pdfauthor={Blake Cole, Gabriel Schamberg},
pdfkeywords={vessel tracking, collision avoidance, unscented Kalman filter, geodetic coordinates, autonomy, AIS},
}

\begin{document}
\maketitle

\begin{abstract}
This paper describes a novel tracking filter, designed primarily for use in collision avoidance systems on autonomous surface vehicles (ASVs).  The proposed methodology leverages real-time kinematic information broadcast via the Automatic Information System (AIS) messaging protocol, in order to estimate the position, speed, and heading of nearby cooperative targets.  The state of each target is recursively estimated in geodetic coordinates using an unscented Kalman filter (UKF) with kinematic equations derived from the spherical law of cosines.  This improves upon previous approaches, many of which employ the extended Kalman filter (EKF), and thus require the specification of a local planar coordinate frame, in order to describe the state kinematics in an easily differentiable form.  The proposed geodetic UKF obviates the need for this local plane.  This feature is particularly advantageous for long-range ASVs, which must otherwise periodically redefine a new local plane to curtail linearization error.  In real-world operations, this recurring redefinition can introduce error and complicate mission planning.  It is shown through both simulation and field testing that the proposed geodetic UKF performs as well as, or better than, the traditional plane-Cartesian EKF, both in terms of estimation error and stability.
\end{abstract}


\section{Introduction}
Collision avoidance is a vital capability of any marine vessel navigating in public waterways; this is particularly true for autonomous surface vehicles (ASVs), which cannot benefit by the real-time guidance of a human operator.  Safe maritime navigation remains a challenge due to the fact that it requires the seamless coordination of multiple complex subsystems.  First, vessels must be able to perceive their surroundings under a wide range of environmental conditions.  This is typically accomplished using one or more \textit{line-of-sight} sensors, which emit electromagnetic or acoustic signals, and detect the reflections produced by nearby obstacles \citep{robinette2019}.  However, in the marine environment, vessels can also utilize the \textit{Automatic Information System} (AIS) protocol to track nearby vessels.  The merits and drawbacks of these sensing modalities will be discussed in Section \ref{ssec:sensing}.  Once an obstacle is detected, the ASV must react quickly and intelligently to avoid it, in accordance with the ``rules of the road'' set forth by the 1972 International Regulations for Prevention of Collisions at Sea (COLREGs) \citep{colregs1972}.  Many ASVs remain unable to perform one or more of these crucial tasks, limiting their adoption beyond the oceanographic research community.

\subsection{Vessel Sensing}
\label{ssec:sensing}
Line-of-sight sensors offer a variety of notable advantages.  Perhaps foremost is their ability to identify \textit{non-cooperative} targets, so named because they do not broadcast their instantaneous kinematic state over designated very-high frequency (VHF) radio channels \citep{lacher2007}.  This is a useful feature, as it permits the detection of both non-transponding vehicles, and non-vehicular obstacles.  However, as their name suggests, line-of-sight sensors also require a direct, unobstructed path between sensor and target, and are thus subject to occlusion by landmasses, fixed infrastructure, and other vessels.  Vessel occlusion is the proximate cause of many collisions at sea, and has received considerable attention in scientific literature \citep{banu2019}.  Furthermore, sensor performance is contingent upon favorable environmental conditions; this is particularly true of LIDAR systems, which are ineffective in particulate-laden air and fog.  RADAR is a highly effective means of marine obstacle detection, but its efficacy, both in terms of angular resolution and range, is dependent upon antenna size and available power (2-5 kW is typical).  Accordingly it is often not a viable option for smaller, power-constrained ASVs.  Further, effective RADAR-based target detection requires careful range-calibration; inattention to this detail has led to tragic collisions of crewed vessels \citep{miller2019}.  While line-of-sight sensors play an invaluable role in marine obstacle detection, they are not without their shortcomings.

\begin{figure}[ht]
    \centering
    \includegraphics[width=0.5\textwidth]{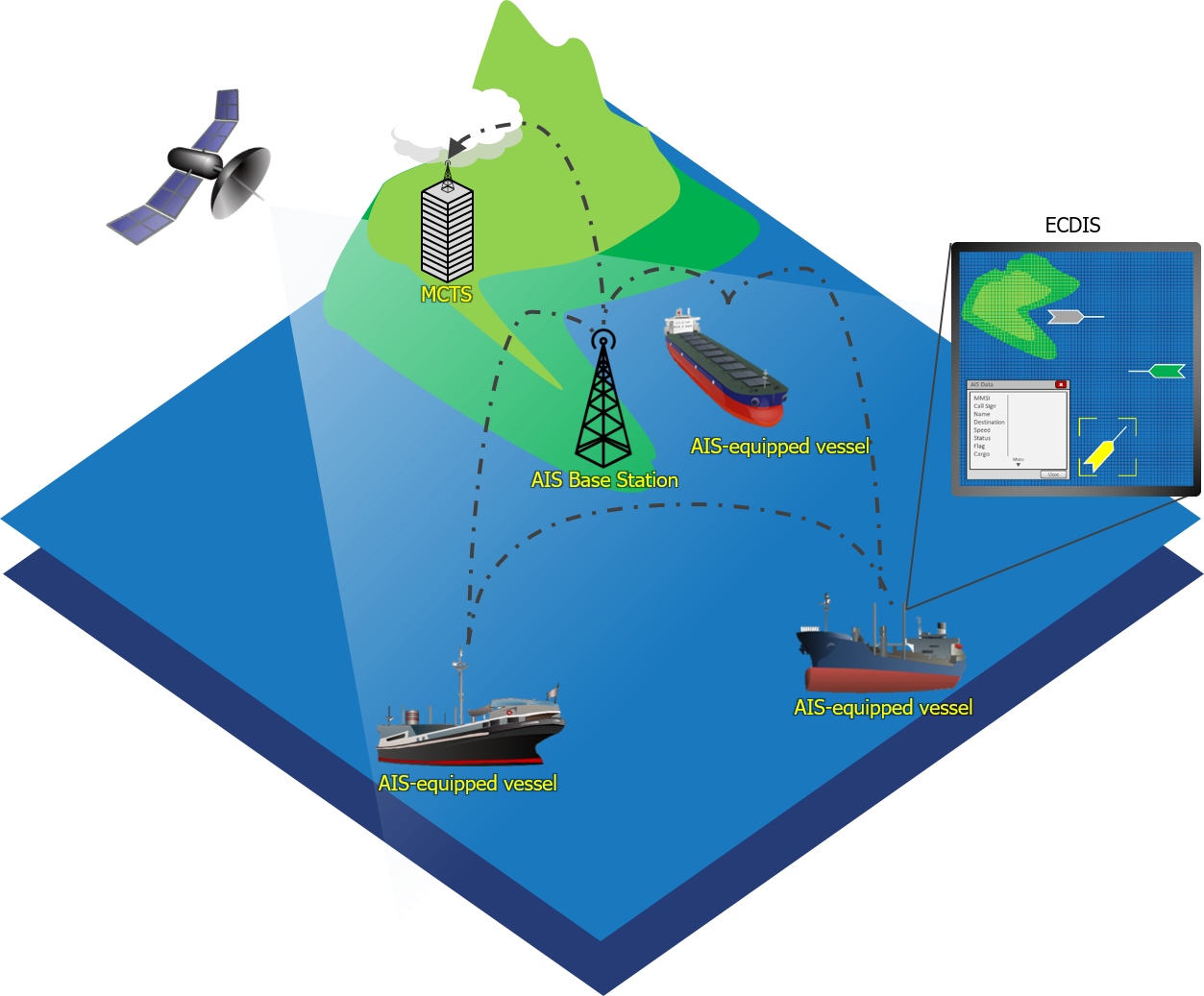}
    \caption{Overview of the Automatic Information System (AIS) \citep{ais_figure}.  AIS allows vessels to share their position, speed, and heading directly with one another.  Some satellites are also capable of detecting AIS messages.  Heavily trafficked areas such as ports and harbors often have one or more terrestrial base stations that act as relays, extending the effective range of AIS transmissions.}
    \label{fig:ais_overview}
\end{figure}

Conveniently, in the marine setting, virtually all commercial vessels are \textit{cooperative}.  The International Maritime Organization’s (IMO) International Convention for the Safety of Life at Sea (SOLAS) requires operating AIS transmitters on all international cargo vessels of more than 300 tons displacement, all cargo vessels of more than 500 tons displacement, and all commercial passenger vessels \citep{imo2002}.  AIS is a VHF radio messaging protocol which permits continuous vehicle-to-vehicle (V2V) communication of key kinematic states, including geodetic position (latitude and longitude), speed over ground (SOG), and course over ground (COG).  The AIS network also includes base stations and –- more recently –- satellites, as illustrated in Fig.  \ref{fig:ais_overview}, to extend the range over which vessels can be tracked.  AIS has a number of considerable advantages over line-of-sight sensors, owing mostly to the particular propagation characteristics of VHF radio waves.  VHF signals can propagate over a considerable range, typically on the order of ten to thirty kilometers, and require only a small amount of power to transmit.  Moreover, they are not particularly sensitive to environmental conditions, and can -- to a limited extent -- circumvent obstacles.  Unlike the data produced by line-of-sight sensors, AIS data is tied to a particular vessel. Each AIS message contains a unique nine-digit identifier, known as the Maritime Mobile Service Identity (MMSI), which corresponds to a single vessel.  This eliminates the need for expensive data association algorithms.

The AIS protocol was originally designed with collision avoidance in mind, and it has been used in this capacity on crewed vessels for decades; however, most ASVs do not leverage this time-tested asset for the purposes of automated collision avoidance.  This is due in part to the sparse transmission interval of AIS messages: ASVs need to make navigation decisions many times per second, and AIS reports are only published once every five to thirty seconds.  The present work describes a novel algorithmic solution to this problem.  If leveraged appropriately, AIS can provide invaluable situational awareness to an otherwise blind ASV, enabling effective collision avoidance behaviors at a marginal cost \citep{cole2021}.  Indeed, an AIS receiver and VHF antenna can be purchased off-the-shelf for less than \$100 USD, whereas most line-of-sight sensors cost thousands to tens of thousands of dollars.

\subsection{Vessel Tracking \& State Estimation}
AIS-based ship traffic prediction has received significant attention in the literature.  Many of these methods rely upon various offline machine learning techniques, including: nearest-neighbor (k-NN) clustering \citep{mazzarella2015a}, traditional neural networks \citep{daranda2016}, convolutional neural networks \citep{kim2018}, recurrent neural networks \citep{gao2018}, and support vector machines \citep{kim2015}.  The primary advantage of such approaches is their ability to generate distant forecasts, on the order of hours or days, by extracting common shipping trajectories from data, and associating individual vessels with those trajectories \citep{pallotta2013}.  This makes them extremely useful for congestion forecasting and traffic management in and around ports.  These methods have also been used in a maritime surveillance context to detect statistically anomalous vessel activity \citep{bomberger2006, ristic2008}.  However, they do not generalize beyond the specific locale in which they were trained.  Further, they are not well-suited for tracking smaller vessels that do not follow predictable trajectories.  These limitations inhibit the application of these methods to short-term vessel motion prediction and collision avoidance.

Short-term, location-independent state estimation and trajectory prediction can be accomplished using a recursive \textit{discretized continuous-time kinematic filter}.  The benefit of this class of filter is two-fold. First, they are able to efficiently smooth noisy measurements in real-time by leveraging data fusion techniques.  Although the precision afforded by modern GNSS systems is generally more than adequate, this is not the only source of error one must consider \citep{matias2015}.  On larger vessels, GNSS receivers are often mounted a considerable distance from the vessel center of flotation, leading to erroneous lateral excursions from the mean trajectory, due to roll and pitch motions.  These spurious oscillations can be mitigated using a properly calibrated kinematic filter.  The second advantage of recursive kinematic filters is that they are able to leverage a process model to provide target state estimates in between successive sensor updates, enabling more effective action-selection on autonomous platforms.  In maritime navigation scenarios, a few seconds can make the difference between a near-miss and a collision.  Recursive kinematic filters can effectively ``fill the gaps'' between sparse sensor updates, affording the ASV autopilot a more accurate sense of its surroundings, and a statistical measure of target position uncertainty.

One of the most popular recursive filters is the discrete linear Kalman filter.  Inconveniently, the linear Kalman Filter (LKF) cannot be applied to AIS-based tracking problems, due to the inherently nonlinear kinematics of vessel motion on an ellipsoidal shell.  \cite{mazzarella2015b} employed a particle filter based approach to rectify this problem, which bears some similarity to the method proposed here.  However, the kinematic equations and their underlying assumptions were not rigorously justified.  Moreover, such approaches are known to be computationally intensive, and unnecessary for mildly nonlinear systems.  The extended Kalman filter (EKF), a sub-optimal variant of the Kalman filter for weakly nonlinear systems, has been used for vessel tracking applications \citep{perera2010, jaskolski2017, fossen2018a}.  \cite{fossen2018b} developed an EKF-variant named the eXogeneous Kalman Filter (XKF) to ameliorate the stability issues associated with the EKF approach.  EKF approaches generally require the projection of geodetic position data onto a local planar coordinate system, in order to describe the system kinematics in a readily differentiable form.  This requirement can prove problematic for a number of reasons.  For one, most satellite navigation systems report position in geodetic coordinates; thus, it is both more natural -- and more computationally efficient -- to track targets directly in this coordinate frame.  Tracking targets in a local plane-Cartesian coordinate system can make the operating region less interpretable, as all target positions are defined relative to an arbitrary origin.  Finally, state estimation accuracy degrades with increasing distance from the local origin; accordingly, ASVs must be programmed to periodically relocate the local coordinate frame to curtail linearization error.  This can complicate mission analysis, and make it difficult to ensure repeatability and consistency across missions.

This manuscript describes a specific instantiation of the unscented Kalman filter (UKF) which obviates the need for local planar coordinate systems by enabling target tracking directly in geodetic coordinates.  This capability stems from a unique feature of the UKF; namely, that it does not rely on a linearized process model to compute the \textit{a priori} state covariance, and thus does not require a Jacobian matrix.  Other authors have used the UKF for vessel tracking, but their filter architectures still require the definition of a local planar coordinate system \citep{duan2008, braca2012}.  The \textit{geodetic UKF} is the primary contribution of this work, and is described in Section \ref{sec:ukf}.  In Section \ref{sec:sim}, the performance of the proposed filter is assessed by way of multiple computer simulations, and it is shown that the filter reliably matches or outperforms a comparable EKF.  Finally, in Section \ref{sec:field}, the geodetic UKF is validated using real AIS data, collected near the entrance to Boston Harbor, USA.  Again, the filter is shown to meet or exceed the performance of a comparable EKF, indicating consistent performance and stability across a wide range of operating conditions.

\section{The Geodetic Unscented Kalman Filter}
\label{sec:ukf}

This section describes the key features of the proposed geodetic UKF.  Section \ref{ssec:kalman_background} gives a brief historical background of the linear Kalman filter and its sub-optimal extensions.  Shortcomings of earlier Kalman filters are discussed in the context of surface vessel tracking.  Section \ref{ssec:implementation} describes the theoretical UKF state estimation approach, and the steps required to implement it.  Section \ref{ssec:ukf_model_derivation} derives the spherical state update equations, and Section \ref{ssec:ukf_sph_approx} investigates their accuracy when applied to the WGS84 Earth ellipsoid.  Sections \ref{ssec:ukf_R} and \ref{ssec:ukf_Q} describe a unique methodology for defining the measurement and process noise covariance matrices.  Lastly, Section \ref{ssec:ukf_angular_residual} covers an important implementation consideration regarding angular residual error.

\subsection{Historical Perspective}
\label{ssec:kalman_background}
The Kalman filter has proven the gold-standard of state estimation since it was first formulated by Rudolf E. Kalman in 1960 \citep{kalman1960}.  Since then, the filter has become ubiquitous, quietly running in the background of smart phones and submarines alike.  The common thread connecting these seemingly disparate use cases is the need for effective sensor fusion: the ability to combine noisy observations with an imperfect process model, yielding a more reliable state estimate (or, at the very least, a statistical measure of the filter's reliability).  The Kalman filter performs this duty admirably.  It guarantees both optimality with respect to state covariance minimization, and uniform asymptotic stability for linear systems with Gaussian noise \citep{gelb1974}.  This is accomplished by solving a discrete variant of the Matrix Riccati Differential Equation, which converges to the Algebraic Riccati Equation (ARE) for observable systems.  For vehicle tracking applications, the primary states of interest generally include: position, orientation, velocity, and occasionally linear or angular acceleration.

\begin{figure}[ht]
    \centering
    \includegraphics[width=0.55\textwidth]{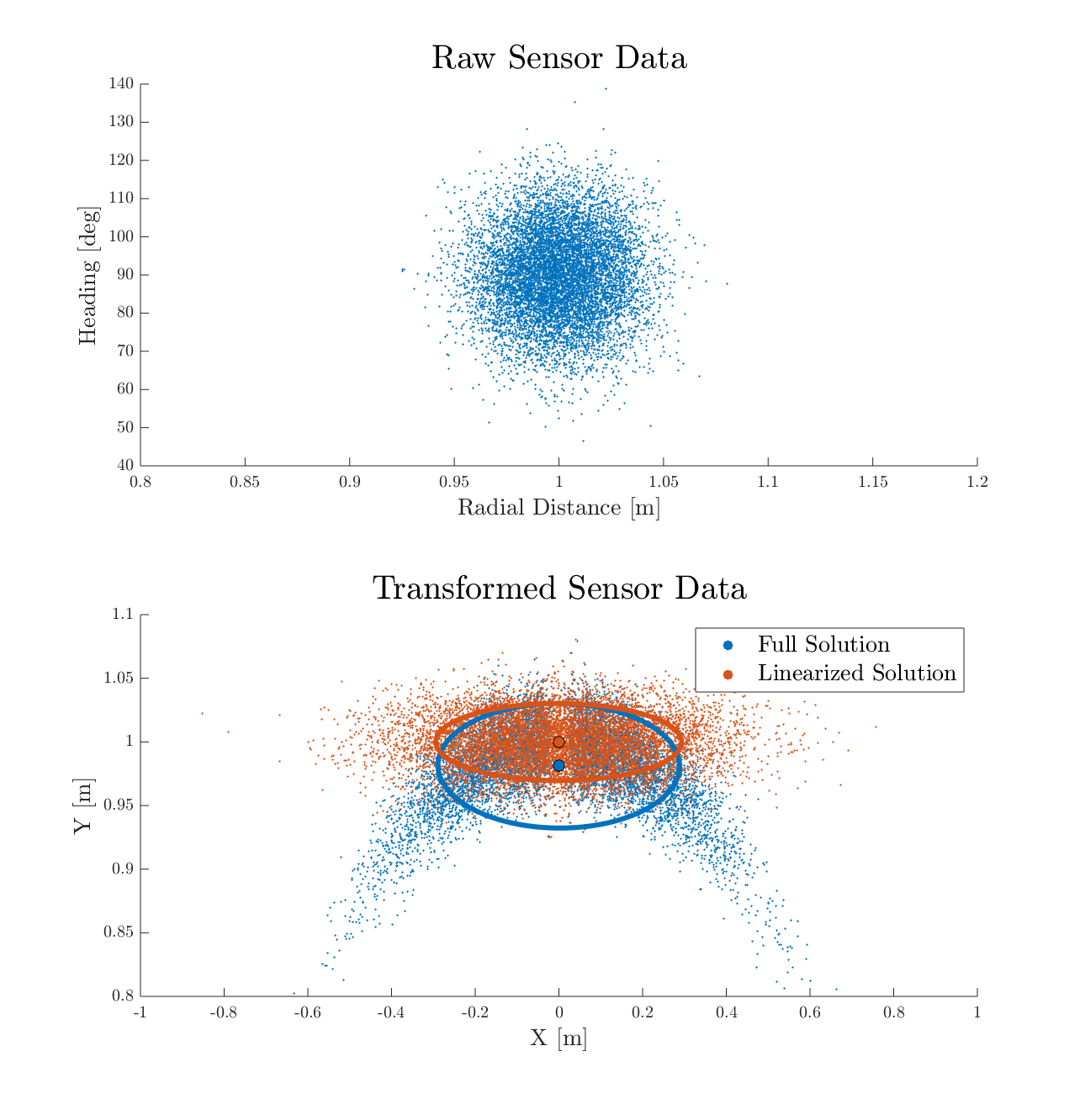}
    \caption{Monte Carlo simulation illustrating the error induced by linearization of a polar-to-Cartesian coordinate transformation, both in terms of expected position and covariance.  This scenario often arises when robotic platforms use line-of-sight sensors for localization.}
    \label{fig:nonlin_transform}
\end{figure}

Unfortunately, many real world systems are either substantially nonlinear, or employ sensors which require nonlinear coordinate transformations.  Take, for example, a line-of-sight sensor which measures the range and bearing to nearby obstacles.  Such a sensor might produce data which resembles the upper subplot of Fig. \ref{fig:nonlin_transform}.  This point cloud can be described by a two-dimensional random variable, with some mean and covariance, described in polar coordinates.  However, expressing this same random variable in Cartesian coordinates produces a warped, non-Gaussian distribution, as shown in the lower subplot of Fig. \ref{fig:nonlin_transform}.  This warping occurs because the the polar-to-Cartesian transformation is nonlinear \citep{braca2012}.  Nonlinear transformations violate the assumptions of the linear Kalman filter, and are incompatible therewith.  Some nonlinear transformations -- including the polar-to-Cartesian transformation -- can be linearized, though not without incurring error, as illustrated by the red point-cloud in the lower subplot of Fig. \ref{fig:nonlin_transform}.  The solid red dot shows the mean position of the linearized solution, whereas the solid blue dot shows the true mean of the resulting distribution.  This error can become very large if there is a large initial variance (i.e. uncertainty) in the azimuthal dimension of the polar random variable.

A maneuvering vessel, tracked in an ellipsoidal reference frame, is a fundamentally nonlinear system.  Fortunately, a variety of suboptimal Kalman filters have been developed to cope with nonlinearity; of these, the extended Kalman filter is likely the most popular and widespread.  According to a NASA technical memorandum published in 1985, this adaptation first grew out of the need to predict orbital trajectories for the Apollo space missions in 1961 \citep{mcgee1985}.  The EKF linearizes the system dynamics about the previous state estimate, and proceeds with the subsequent state prediction and correction steps, as though the system were linear.  This strategy has a few drawbacks.  First and foremost, the EKF is known to underestimate covariance.  This issue is sometimes ameliorated by adding so-called ``stabilizing noise'' in an \textit{ad hoc} manner, but this does not guarantee stability.  Because the filter linearizes the system about a potentially inaccurate previous state estimate, the EKF offers only local stability at best, and can quickly diverge when applied to highly nonlinear systems with inadequate temporal resolution.  Further, it requires a Jacobian matrix, which may be prohibitively expensive to compute for high-dimensional systems, or systems without analytical derivatives.

In the late 1990s, a new variant of Kalman filter called the unscented Kalman filter was developed to address these shortcomings \citep{julier1997}.  Unlike the EKF, the UKF does not require a Jacobian matrix; rather, it employs an unscented transform to propagate the system state and covariance from one discrete time to the next.  This approach is justified by the contention that it is easier to approximate the statistical moments of a non-Gaussian distribution than it is to linearize a nonlinear transformation \citep{julier2004}.  The UKF bears aesthetic similarities to a particle filter, with the key difference being that the inputs are not chosen at random; rather, the so-called \textit{sigma points} are deterministically chosen to preserve higher-order information about the distribution of a random variable, using a small number of points.  Consequently, the computational cost of the UKF is much lower than that of a particle filter, and similar to that of an EKF.  Further, it is understood that the UKF outperforms the EKF for highly-nonlinear systems, or systems with rapidly varying covariance.  Indeed, methodologies for UKF sigma point selection have been developed such that third-order accuracy is guaranteed for Gaussian inputs, for all nonlinearities \citep{wan2000}.  Even for non-Gaussian inputs, the UKF can always achieve at least second-order accuracy, while the EKF guarantees only first-order accuracy \citep{wan2000}.  Finally, for the particular application at hand -- target tracking in geodetic coordinates -- the EKF generally requires projection of geodetic position data onto a local planar coordinate frame, in order to describe the system kinematics in an easily differentiable form.  This linearization technique can incur substantial error when used to estimate the position of targets located far from the local origin.  This source of error is characterized in Appendix \ref{appendix:plane_error}.

\subsection{Implementation}
\label{ssec:implementation}
For AIS-based tracking applications, the use of a geodetic process model enables the use of a simple linear observer.  Accordingly, while the state and covariance matrix must be \textit{predicted} using the UKF approach, they can be \textit{updated} using the conventional linear approach.

Assume the system can be characterized by a state vector $\boldsymbol{x}_k$ = [$lon_k$, $lat_k$, $U_k$, $\alpha_k$], with dimension $N_x = 4$, at discrete time $k$.  The individual elements of the state vector are: longitude ($lon_k$), latitude ($lat_k$), SOG ($U_k$), and COG ($\alpha_k$).  The system dynamics are given by the vector of nonlinear, geodetic state update functions $\boldsymbol{f}(\boldsymbol{x})$, which are derived in Section \ref{ssec:ukf_model_derivation}.  This process model enables the use of a simple linear observer $\boldsymbol{H} = \boldsymbol{I}$.  Both the process and measurement models are affected by additive, zero-mean Gaussian noise, represented in Eqs. \ref{eq:generic_state_update} and \ref{eq:generic_observer} as random variables $\boldsymbol{w}_k$ and $\boldsymbol{v}_k$, respectively ($\E[\boldsymbol{w}_k] = \E[\boldsymbol{v}_k] = 0$ for all $k$).

\begin{align}
    \quad \boldsymbol{x}_{k+1} &= \boldsymbol{f}(\boldsymbol{x}_k) + \boldsymbol{w}_k, \qquad\qquad\!  \boldsymbol{w}_k \sim \mathcal{N}(0, \boldsymbol{Q}_k) && \label{eq:generic_state_update}\\[4pt]
    \quad \boldsymbol{z}_k &= \boldsymbol{H}\boldsymbol{x}_k + \boldsymbol{v}_k, \qquad\qquad\;\;  \boldsymbol{v}_k \sim \mathcal{N}(0, \boldsymbol{R}_k)  \label{eq:generic_observer}
\end{align}

The process noise covariance matrix $\boldsymbol{Q}_k$ is assumed to be positive semi-definite.  This ensures that the Gaussian distribution described by $\boldsymbol{Q}_k$ is well-defined.  Because there is no such thing as a perfect sensor, the measurement noise covariance matrix $\boldsymbol{R}_k$ is assumed to be positive-definite (though it can be very small).  It is further assumed that $\boldsymbol{w}_k$ and $\boldsymbol{v}_k$ are uncorrelated at all times $k$ and $l$, such that:

\begin{align}
    \E[\boldsymbol{w}_k \boldsymbol{w}_l] &= \E[\boldsymbol{v}_k \boldsymbol{v}_l] = 0, \qquad\quad\; k \neq l &&\\[4pt]\nonumber
    \E[\boldsymbol{w}_k \boldsymbol{v}_l] &= 0 &&\nonumber
\end{align}

Like all Kalman Filters, the UKF employs the \textit{predictor-corrector} paradigm.  The most recent \textit{a posteriori} state and covariance estimate is used in conjunction with a process model to predict the \textit{a priori} state and covariance at some future time.  The moment a measurement is available, it is compared with the prediction, and the two estimates are fused to produce a more reliable state estimate.

\bigskip
\begin{enumerate}
\item Predict
    \begin{enumerate}
        \item \textit{Use the prior state estimate and covariance matrix to compute sigma points and weights.}  For typical vessel maneuvers and filter update rates, the geodetic state update equations are expected to exhibit only mild nonlinearity.  Accordingly, the unscented transform should produce a distribution with little or no skew.  For such systems, it is best to use a \textit{symmetric} sigma point selection scheme \citep{julier2004}.  We use $2N_x + 1$ sigma points $\mathcal{X}^{(i)}$, with corresponding weights $W^{(i)}$, as suggested by Julier and Uhlmann \citep{julier2004}.  This set of $N_x$-dimensional points are redefined at each discrete time $k$.  The \textit{a posteriori} mean and covariance are given by $\boldsymbol{\Hat{x}_k}$ and $\boldsymbol{P_k}$, respectively.
        
        \begin{align}
            \mathcal{X}_k^{(0)} &= \boldsymbol{\Hat{x}}_{k-1} &&\\[4pt]\nonumber
            W^{(0)} &= 1 - \frac{N_x}{3} &&\\[4pt]\nonumber
            \mathcal{X}_k^{(i)} &= \boldsymbol{\Hat{x}}_{k-1} + \left( \sqrt{\frac{N_x}{1-W^{(0)}} \boldsymbol{P}_{k-1}} \right)_i, \qquad i = 1,2,...,N_x&&\\[4pt]\nonumber
            \mathcal{X}_k^{(i+N_x)} &= \boldsymbol{\Hat{x}}_{k-1} - \left( \sqrt{\frac{N_x}{1-W^{(0)}} \boldsymbol{P}_{k-1}} \right)_i, \qquad i = 1,2,...,N_x&&\\[4pt]\nonumber
            W^{(i)} &= \frac{1 - W^{(0)}}{2N_x}, \qquad\qquad\qquad\qquad\qquad\; i = 1, 2,...,2N_x&&\\\nonumber
        \end{align}
        
        \item \textit{Pass sigma points through nonlinear transformations.}  There is quite a bit of flexibility in this step.  The user has the option to chose one of various Earth models (i.e. spherical, WGS84), and can easily implement various maneuvering models by adjusting scalar constants in the SOG and COG update equations.  If an ellipsoidal Earth model is chosen, the position update equations must solved using an iterative method like Vincenty's Formulae \citep{vincenty1975}.  The use of an ellipsoidal Earth model yields more accurate state estimates than can be attained using a spherical Earth model, but incurs substantial computational overhead, and is shown in Section \ref{ssec:ukf_sph_approx} to be unnecessary for most practical vessel tracking applications.  In Eq. \eqref{eq:transformed points}, the sigma points $\mathcal{X}^{(i)}$ are passed through a chosen nonlinear transformation $\boldsymbol{g}$, yielding the transformed set of $N_x$-dimensional points $\mathcal{Y}^{(i)}$.
        
        \begin{equation}
            \label{eq:transformed points}
            \mathcal{Y}_k^{(i)} = \boldsymbol{g}(\mathcal{X}_k^{(i)}) \qquad\qquad i = 0, 1,...,2N_x
        \end{equation}
        
        \item \textit{Compute the \textit{a priori} state estimate and covariance matrix using the weighted, transformed sigma points.}  The nonlinear transformation of sigma points generally yields a non-Gaussian distribution in $\mathbb{R}^{N_x}$ space.  The efficacy of the UKF approach hinges upon the fundamental assumption that it is reasonable to treat this distribution as though it were approximately Gaussian.  This hypothesis has been widely tested, and holds true in many practical applications.
        
        \begin{align}
            \boldsymbol{\Hat{x}}_{k|k-1} &= \sum_{i=0}^{2N_x} W^{(i)} \mathcal{Y}_k^{(i)} && \label{eq:apriori_x}\\[4pt]
            \boldsymbol{P}_{k|k-1} &= \sum_{i=0}^{2N_x} W^{(i)} \{\mathcal{Y}_k^{(i)} - \boldsymbol{\Hat{x}}_{k|k-1}\}\{\mathcal{Y}_k^{(i)} - \boldsymbol{\Hat{x}}_{k|k-1}\}^T + \boldsymbol{Q}_k \label{eq:apriori_P}
        \end{align}
    \end{enumerate}

\medskip
\item Update
    \begin{enumerate}
        \item \textit{Compute the Kalman gain using the \textit{a priori} covariance.}
        
        \begin{equation}
            \boldsymbol{K}_k = \boldsymbol{P}_{k|k-1} \boldsymbol{H}^T \left[ \boldsymbol{H} \boldsymbol{P}_{k|k-1} \boldsymbol{H}^T + \boldsymbol{R}_k \right]^{-1}
        \end{equation}
        
        \item \textit{Map the state prediction into the measurement space, and compute the residual error.}  It is not uncommon for AIS reports to contain missing fields.  Whenever this is the case, the missing element $z_k^j$ in the measurement vector $\boldsymbol{z}_k$, and its corresponding diagonal element in $\boldsymbol{H}$, must both be set equal to zero.  This eliminates the residual error for the unobserved state only, indicating to the filter that it should rely exclusively upon the \textit{a priori} state estimate to predict $\Hat{x}_k^j$ at time $k$.  Meanwhile, the other states in $\boldsymbol{\Hat{x}}_{k|k-1}$ are fused with valid measurements as usual.
        
        \begin{equation}
            \boldsymbol{y}_k = \boldsymbol{z}_k - \boldsymbol{H} \boldsymbol{\Hat{x}}_{k|k-1}
        \end{equation}

        \item \textit{Compute the \textit{a posteriori} state estimate and covariance matrix.}  The numerically stable Joseph form of the covariance update equation is used to prevent the loss of matrix symmetry due to floating point errors \citep{gelb1974}.
        
        \begin{align}
            \boldsymbol{\Hat{x}}_k &= \boldsymbol{\Hat{x}}_{k|k-1} + \boldsymbol{K}_k \boldsymbol{y}_k &&\\[4pt]
            \boldsymbol{P}_k &= \left[\boldsymbol{I} - \boldsymbol{K}_k \boldsymbol{H} \right] \boldsymbol{P}_{k|k-1} \left[\boldsymbol{I} - \boldsymbol{K}_k \boldsymbol{H} \right]^T + \boldsymbol{K}_k \boldsymbol{R}_k \boldsymbol{K}_k^T
        \end{align}
    \end{enumerate}
\end{enumerate}

The statistical properties and theoretical justification of the UKF will not be covered here; many exemplary resources covering this topic already exist.  The reader is directed to \cite{labbe2005} for an introductory text; and to \cite{julier2004} and \cite{wan2000} for more a more advanced theoretical treatment of the topic.

\subsection{Process Model Derivation}
\label{ssec:ukf_model_derivation}
For the purposes of the proposed vessel-tracking approach, the most important feature of the UKF is its compatibility with nonlinear systems with cumbersome derivatives.  While it is not impossible to compute the Jacobian matrix of a geodetic process model comprised of Eqs. \eqref{eq:latupdate2}, \eqref{eq:lonupdate3}, \eqref{eq:SOGupdate}, and \eqref{eq:COGupdate}, the resulting analytical derivatives are unwieldy and involve hundreds of nonlinear terms.  The UKF is a better choice for such complicated nonlinear systems.  This approach permits the description of the system kinematics in the same coordinate frame as the sensor (AIS) input.  A static variant of the proposed kinematic state update equations can be found in a 1987 United States Geological Survey technical report \citep{snyder1987}; however, to the best of the authors' knowledge, no comprehensive proof of this relationship currently exists in the academic or technical literature.  Accordingly, one is provided here for completeness.
 
Any plane coincident with the center point of a sphere defines a great circle path upon its surface, named so because this path forms the largest circle which can be drawn upon the surface of a sphere. This particular line is of great interest to navigators, for the shortest path between any two points on the surface of a sphere can be described by a unique great circle path (with the exception of antipodal points, which can be joined by an infinite number of great circle paths).  A \textit{spherical triangle} is formed by the intersection of three such paths, and obeys a number of laws derived from planar geometry relationships.  These laws can be exploited for the purposes of spherical navigation, and -- combined with the principles of classical mechanics -- accurately describe discrete spherical kinematics.

\begin{figure}[ht]
    \centering
    \includegraphics[width=0.3\textwidth]{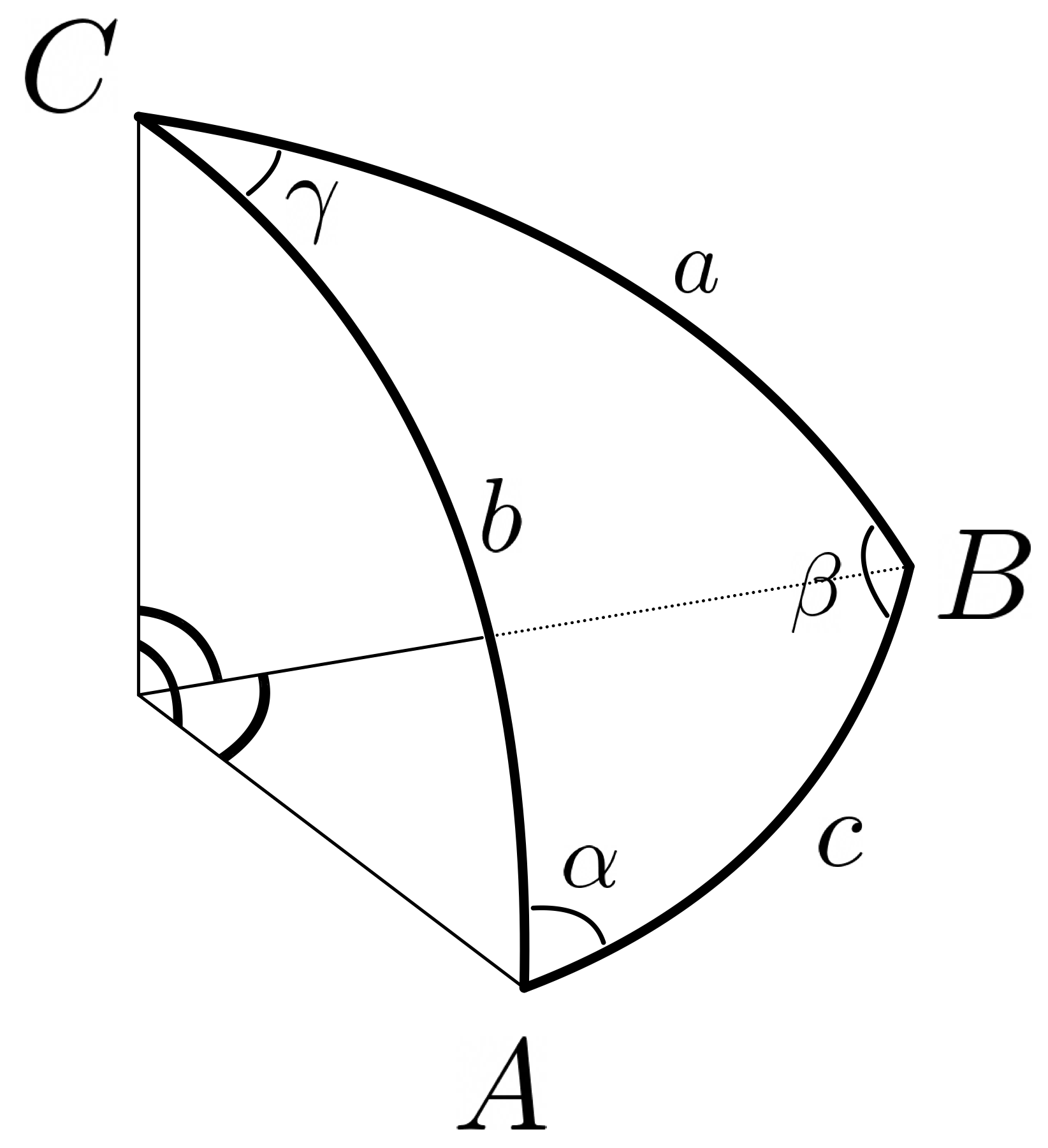}
    \caption{Unit spherical triangle.  Vertices are denoted by uppercase letters; sector angles by lowercase letters; and internal angles by lowercase Greek letters $\alpha$, $\beta$, and $\gamma$.}
    \label{fig:sph_triangle}
\end{figure}

\subsubsection{Latitude}
\label{sssec:lat}
The spherical law of cosines was originally derived to express an internal angle of a spherical triangle in terms of the length of its three sides (expressed as sector angles) \citep{todhunter1886}.  A unit-spherical triangle is shown in Fig. \ref{fig:sph_triangle}.  The canonical law of cosines states:

\begin{equation}
    \label{eq:loc1}
    \cos{(c)} = \cos{(a)}\cos{(b)} + \sin{(a)}\sin{(b)}\cos{(\gamma)}
\end{equation}

For a thorough derivation and proof, the reader is directed to trigonometry textbooks by \cite{todhunter1886}, or \cite{bocher1914}.  Cyclic permutation of Eq. \eqref{eq:loc1} gives:

\begin{equation}
    \label{eq:loc2}
    \cos{(a)} = \cos{(b)}\cos{(c)} + \sin{(b)}\sin{(c)}\cos{(\alpha)}
\end{equation}

If the sphere in question is assumed to be Earth, and the point $C$ is fixed at the North Pole, the angle $\gamma$ becomes the difference in longitude between points $A$ and $B$.  The angles $a$ and $b$, which subtend the meridional arcs joining the North Pole and points $A$ and $B$, respectively, and represent the complement of the latitude of points $A$ and $B$, sometimes referred to as the \textit{co-latitudes}.

\begin{align} 
    \cos{(a}) &= \cos{(90^{\circ}-lat_B)} = \sin{(lat_B)} \\[4pt]
    \cos{(b}) &= \cos{(90^{\circ}-lat_A)} = \sin{(lat_A)}
\end{align}

Unless explicitly stated otherwise, assume that any trigonometric function applied to latitude ($lat$), longitude ($lon$), or departure angle ($\alpha$) is converted to radians prior to evaluation.  The central angle between points $A$ and $B$ can be written \(c=\frac{d}{R}\), where $d$ is the great-circle distance between points $A$ and $B$, and and $R$ is the sphere radius.    These observations permit a useful restatement of Eq. \eqref{eq:loc2} as:

\begin{equation}
    \label{eq:latupdate1}
        \sin{(lat_B)} = \sin{(lat_A)}\cos{(\textstyle\frac{d}{R})} + \cos{(lat_A)}\sin{(\textstyle\frac{d}{R})}\cos{(\alpha)}
\end{equation}

This equation determines the latitude of a new point ($lat_B$) on the surface of a sphere in $\mathbb{R}^3$, given a distance ($d$) traveled along the great circle path defined by the latitude of a prior point ($lat_A$), and an initial departure angle ($\alpha$).  Note that the two points $A$ and $B$, separated a by scalar arc length ($d$), will always lie on the unique line defined by $lat_A$ and $\alpha$.  This subtle observation begets an invaluable simplification: along this line, the three-dimensional spherical kinematics can be described using one-dimensional Newtonian particle mechanics. In other words, the great circle path defined by $lat_A$ and $\alpha$ can be thought of as a level-set of the sphere in $\mathbb{R}^3$, mapping $f : \mathbb{R}^3 \rightarrow \mathbb{R}$.

\begin{equation}
    \label{eq:r3r1}
    \{(x,y,z) \in \mathbb{R}^3 \mid  f(x,y,z) = c_0\}
\end{equation}

Along this line in $\mathbb{R}$, the kinematic equations of motion can be solved easily, giving spatial displacement ($d$) as a function of scalar speed ($v$) and time ($t$):

\begin{align} 
    \label{eq:kinematics}
    v  &= \frac{dx}{dt} \\
    \int_{x_0}^{x} dx &= \int_{t_0}^{t} v dt \\
    x &= x_0 + v \Delta t \label{eq:particle_mechanics}
\end{align}

Eq. \ref{eq:particle_mechanics} describes the distance traveled by a particle moving at a constant speed ($v$), over a fixed time interval ($\Delta t$).  For the purposes of vessel tracking, the distance $v \Delta t$ can be interpreted as the arc length $d$, while $v$ is the previous SOG state estimate ($U_{k-1}$), which is assumed constant over the filter update interval ($\Delta t_k$).  Implementing this result in Eq. \eqref{eq:latupdate1} gives the final discrete state update equation for latitude:

\begin{equation}
    \label{eq:latupdate2}
    \boxed{
            lat_k = \arcsin{\Big[\sin{(lat_{k-1})}\cos{(\textstyle\frac{U_{k-1} \Delta t_k}{R})} + \cos{(lat_{k-1})}\sin{(\textstyle\frac{U_{k-1} \Delta t_k}{R})}\cos{(\alpha_{k-1})}\Big]}
    }
\end{equation}

\subsubsection{Longitude}
\label{sssec:lon}
The relationship derived above reveals that the arc length spanned by one degree of latitude is independent of longitudinal position; however, the converse is not true.  Consequently, the derivation of the longitude update equation is somewhat more involved.  At first glance, it would seem a simple rearrangement of the canonical law of cosines could provide a formula for the change in longitude, captured by the term $\cos{(\gamma)}$.  However, $\arccos{(\gamma)}$ is ill-conditioned for small values of $\gamma$, and thus must be eliminated from the update equation.  This is accomplished using the spherical law of sines, which states:

\vspace*{-\baselineskip} 
\begin{equation}
    \label{eq:los}
    \frac{\sin{(\alpha)}}{\sin{(a)}} = \frac{\sin{(\beta)}}{\sin{(b)}} = \frac{\sin{(\gamma)}}{\sin{(c)}}
\end{equation}
\smallskip

A thorough derivation and proof of this relationship is available in \cite{todhunter1886}, or \cite{bocher1914}. Again, the canonical spherical law of cosines states:

\begin{equation*}
    \tag{\ref{eq:loc1}}
    \cos{(c)} = \cos{(a)}\cos{(b)} + \sin{(a)}\sin{(b)}\cos{(\gamma)}
\end{equation*}

Multiplication of both sides of Eq. \eqref{eq:loc1} by carefully selected equivalent ratios from the spherical law of sines, Eq. \eqref{eq:los}, gives:

\vspace*{-\baselineskip} 
\begin{equation}
    \label{eq:loc_los}
        \left( \frac{\sin{(\gamma)}}{\sin{(c)}} \right) \cos{(c)} = \left( \frac{\sin{(\gamma)}}{\sin{(c)}} \right) \cos{(a)} \cos{(b)} + \left( \frac{\sin{(\alpha)}}{\sin{(a)}} \right) \sin{(a)} \sin{(b)} \cos{(\gamma)}
\end{equation}

Dividing both sides of Eq. \eqref{eq:loc_los} by $\cos{(\gamma)}$, and rearranging slightly, produces an expression for an arbitrary change in longitude between two points on a sphere ($\gamma$), given the co-latitudes of both points ($a$ and $b$), and the angular distance ($c$) travelled in the direction specified by the initial departure angle ($\alpha$).

\begin{gather}
    \tan{(\gamma)} \left( \frac{\cos{(c)}}{\sin{(c)}} \right) =
    \tan{(\gamma)} \left( \frac{\cos{(a)}\cos{(b)}}{\sin{(c)}} \right) + \sin{(b)} \sin{(\alpha)}\\
    \tan{(\gamma)} =
    \tan{(\gamma)} \left( \frac{\cos{(a)}\cos{(b)}}{\cos{(c)}} \right) + \frac{\sin{(b)}\sin{(c)}\sin{(\alpha)}}{\cos{(c)}}\\
    \!\tan{(\gamma)} \left( 1 - \frac{\cos{(a)}\cos{(b)}}{\cos{(c)}} \right) \; = \;
    \frac{\sin{(b)}\sin{(c)}\sin{(\alpha)}}{\cos{(c)}}\\
    \tan{(\gamma)} = \quad \frac{\sin{(b)}\sin{(c)}\sin{(\alpha)}}{\cos{(c)} - \cos{(a)}\cos{(b)}}
    \label{eq:lonupdate1}
\end{gather}

Eq. \eqref{eq:lonupdate1} is a computationally well-conditioned expression for change in longitude ($\gamma$); however, it requires prior knowledge of the co-latitude of both points in question ($a$ and $b$), which precludes its use as a viable position update equation.  Fortunately, this issue can be rectified by substituting a cyclic permutation of Eq. \eqref{eq:loc1} in Eq. \eqref{eq:lonupdate1}:

\begin{align}
    \tan{(\gamma)} &=
    \frac{\sin{(b)}\sin{(c)}\sin{(\alpha)}}{\cos{(c)} - \cos{(b)} \left[\cos{(b)}\cos{(c)} + 
    \sin{(b)}\sin{(c)}\cos{(\alpha)} \right]} \\
    &=
    \frac{\sin{(b)}\sin{(c)}\sin{(\alpha)}}{\cos{(c)} - \cos{(c)}\big[1 - \sin^{2}(b)\big] - \cos{(b)}\sin{(b)}\sin{(c)}\cos{(\alpha)}} \\
    &=
    \frac{\sin{(b)}\sin{(c)}\sin{(\alpha)}}{\cos{(c)}\sin^{2}(b) - \cos{(b)}\sin{(b)}\sin{(c)}\cos{(\alpha)}} \\
    &= \frac{\sin{(c)}\sin{(\alpha)}}{\cos{(c)}\sin{(b)} - \cos{(b)}\sin{(c)}\cos{(\alpha)}}
    \label{eq:lonupdate2}
\end{align}

Expressing the co-latitudes in Eq. \eqref{eq:lonupdate2} as latitudes, and restating the internal angle between points $A$ and $B$ in kinematic terms gives the final discrete state update equation for longitude:

\begin{equation}
    \label{eq:lonupdate3}
    \boxed{
        \begin{aligned}
            lon_k &= lon_{k-1} + \\ &\atantwo \Big[\sin{(\textstyle\frac{U_{k-1} \Delta t_k}{R})} \sin{(\alpha_{k-1})},
            \cos{(lat_{k-1})}\cos{(\textstyle{\frac{U_{k-1} \Delta t_k}{R})}} -
            \sin{(lat_{k-1})}\sin{(\textstyle{\frac{U_{k-1} \Delta t_k}{R})}}\cos{(\alpha_{k-1})}\Big]
        \end{aligned}
    }
\end{equation}

\subsubsection{Speed Over Ground (SOG)}
\label{sssec:SOG}
Linear acceleration can be estimated using a first-order backwards difference approximation, with a variable time step, based on prior observations of vessel speed \citep{fossen2018a}.  This acceleration estimate can then be used in conjunction with a prior speed estimate to form a valid state update equation:

\begin{equation}
    \label{eq:SOGupdate}
    \boxed{ U_k = U_{k-1} + a_{k-1} \Delta t_k}
\end{equation}

Alternatively, $a_k$ can be set equal to zero, in which case a \textit{constant-speed} motion model is recovered.  This assumption is reasonable, and yields a more robust filter in practice.  AIS transmission intervals are almost always in excess of 6 seconds; consequently, computing SOG using a backwards difference approximation can lead to rapid divergence and instability.  Furthermore, as the sensor update interval becomes shorter, the usefulness of a backwards difference approximation diminishes.

\subsubsection{Course Over Ground (COG)}
\label{sssec:COG}
Similarly, turning rate can be estimated using a first-order backwards difference approximation, with a variable time step, based on prior observations of vessel heading:

\begin{equation}
    \label{eq:COGupdate}
    \boxed{\alpha_k = \alpha_{k-1} + r_{k-1} \Delta t_k}
\end{equation}
However, this approach suffers the same robustness issues described in Section \ref{sssec:SOG}.  Accordingly, $r_k$ is set to zero in the present work, producing a \textit{constant-heading} motion model.  When both $a=0$ and $r=0$, a \textit{constant-velocity} (CV) motion model is recovered.  In Sections \ref{sec:sim} and \ref{sec:field}, it is shown by way of simulation and field testing, respectively, that the simple CV motion model produces admirable results when applied to the surface vessel tracking problem.  For high-speed tracking applications, it may be necessary to construct a \textit{multiple model filter}, containing an ensemble of motion models.  The structure of the proposed state equations makes the creation of more complicated maneuvering models a trivial affair.  For instance, a constant acceleration (CA) model can be created by defining a constant acceleration ($a_k = a$); alternatively, a constant turn (CT) model can be created by defining a constant rate of turn ($r_k = r$).  This feature drastically alleviates the burden of creating multiple model filters.

\smallskip
\begin{figure}[ht]
\label{fig_ukf_approach}
\centering
\underline{\textsc{Geodetic UKF State Update Equations (Constant Velocity)}} \\[8pt]
\textsc{\small{Nonlinear Process Model:}}
\begin{equation}
    \label{eq:ukf_model}
    \footnotesize
    \hspace*{-2mm}
    \begin{bmatrix}
        lon \\ lat \\ U \\ \alpha
    \end{bmatrix}_{k}
    =
    \begin{bmatrix}
        lon_{k-1} + \atantwo \big[\sin{(\textstyle\frac{U_{k-1} \Delta t_k}{R})} \sin{(\alpha_{k-1})},\;
        \cos{(lat_{k-1})}\cos{(\textstyle{\frac{U_{k-1} \Delta t_k}{R})}} -
        \sin{(lat_{k-1})}\sin{(\textstyle{\frac{U_{k-1} \Delta t_k}{R})}}\cos{(\alpha_{k-1})}\big]
        \\
        \arcsin\big[\sin{(lat_{k-1})}\cos{(\textstyle\frac{U_{k-1} \Delta t_k}{R})} \; + \;
        \cos{(lat_{k-1})}\sin{(\textstyle\frac{U_{k-1} \Delta t_k}{R})}\cos{(\alpha_{k-1})}\big]
        \\
        U_{k-1} + a_{k-1} \Delta t_k\\
        \alpha_{k-1} + r_{k-1} \Delta t_k      
    \end{bmatrix}
\end{equation}\\[6pt]

\textsc{\small{Linear Measurement Model:}}
\begin{equation}
    \label{eq:lkf_obs}
    \footnotesize
    \begin{bmatrix}
        lon \\ lat \\ U \\ \alpha
    \end{bmatrix}_{meas.}
    =
    \begin{bmatrix}
        1 & 0 & 0 & 0 \\
        0 & 1 & 0 & 0 \\
        0 & 0 & 1 & 0 \\
        0 & 0 & 0 & 1        
    \end{bmatrix}
    \begin{bmatrix}
        lon \\ lat \\ U \\ \alpha
    \end{bmatrix}_{state}
\end{equation}
\caption{Unscented Kalman filter (UKF) update equations and linear measurement model.}
\end{figure}

\subsection{Spherical Approximation Error}
\label{ssec:ukf_sph_approx}
The first two terms of the proposed state update matrix (Eqs. \eqref{eq:latupdate2} and \eqref{eq:lonupdate3}) are derived from fundamental identities of spherical trigonometry; accordingly, they are mathematically exact only for spherical bodies.  Somewhat inconveniently, the earth is not a sphere; rather, it is an oblate ellipsoid (to close approximation), with its polar radius being some 0.34\% shorter than its equatorial radius \citep{nga_2014}.  Accordingly, the spherical state update equations derived in Sec. \ref{ssec:ukf_model_derivation} will always incur some error.  Methods have been devised for calculating geodetic curves on ellipsoids of revolution.  Most notable of these is Vincenty's Formulae, an iterative technique which leverages Legendre polynomials to map an ellipsoidal geodetic curve onto a great circle path on an auxiliary sphere.  This method is accurate to less than 0.5mm on the Earth ellipsoid \citep{vincenty1975}, and thus serves as a useful reference point for comparison.

\begin{figure}[ht]
\label{fig:spherical_approx_error}
    \centering
    \includegraphics[width=0.82\textwidth]{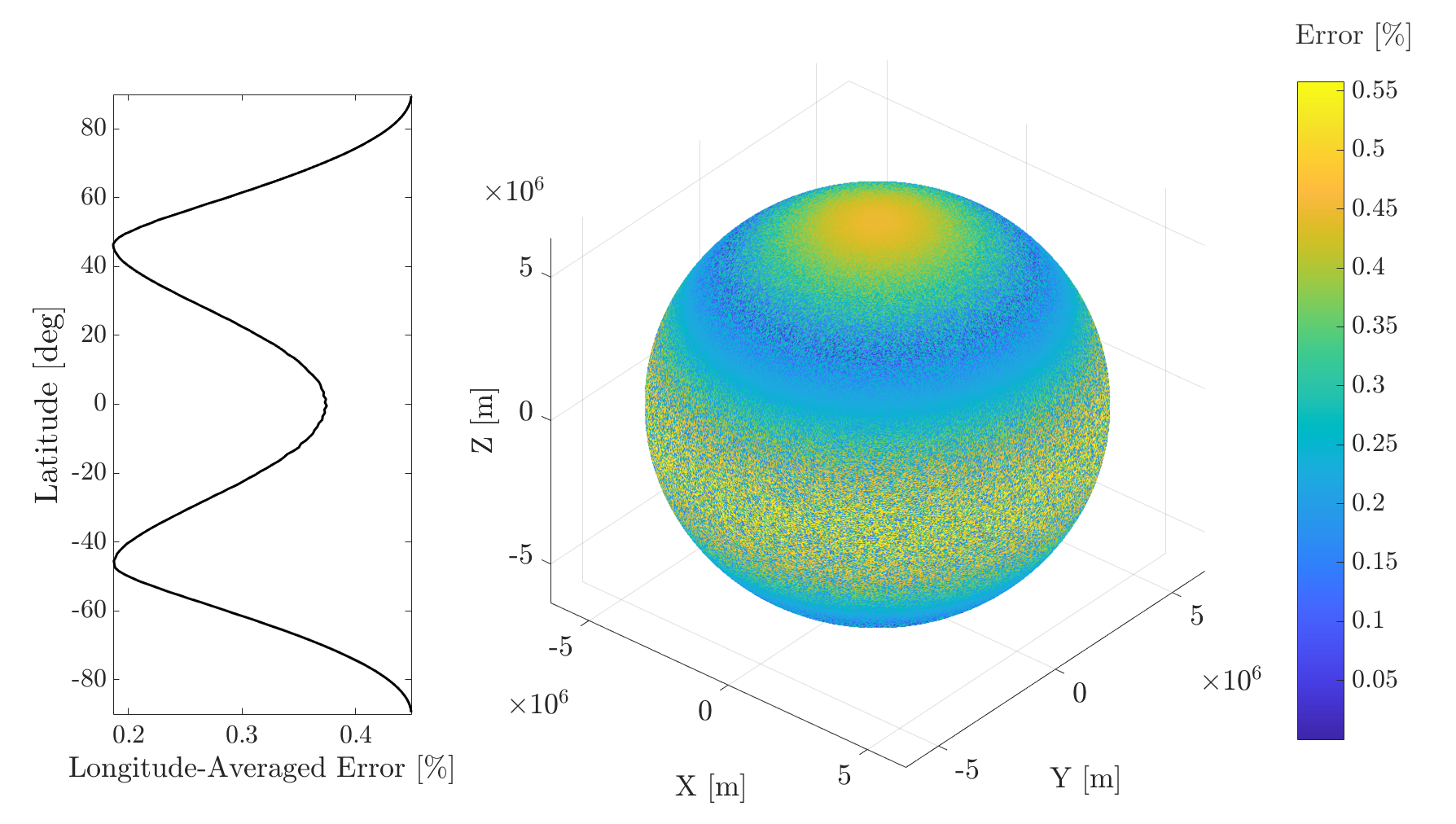}
    \caption{Illustration of the distance-normalized position error incurred by the proposed spherical state update equations, using as ground truth the iterative method proposed by \cite{vincenty1975}, which is known to yield sub-millimeter accuracy on the WGS84 Earth ellipsoid model.  The normalized error is smallest in the mid-latitudes, and grows larger near the poles and equator.  Nevertheless, the spherical approximation provides sub-meter accuracy for arc lengths less than 200m, anywhere on Earth.}
\end{figure}

\vfill
\begin{figure}[ht]
    \centering
    \includegraphics[width=0.48\textwidth]{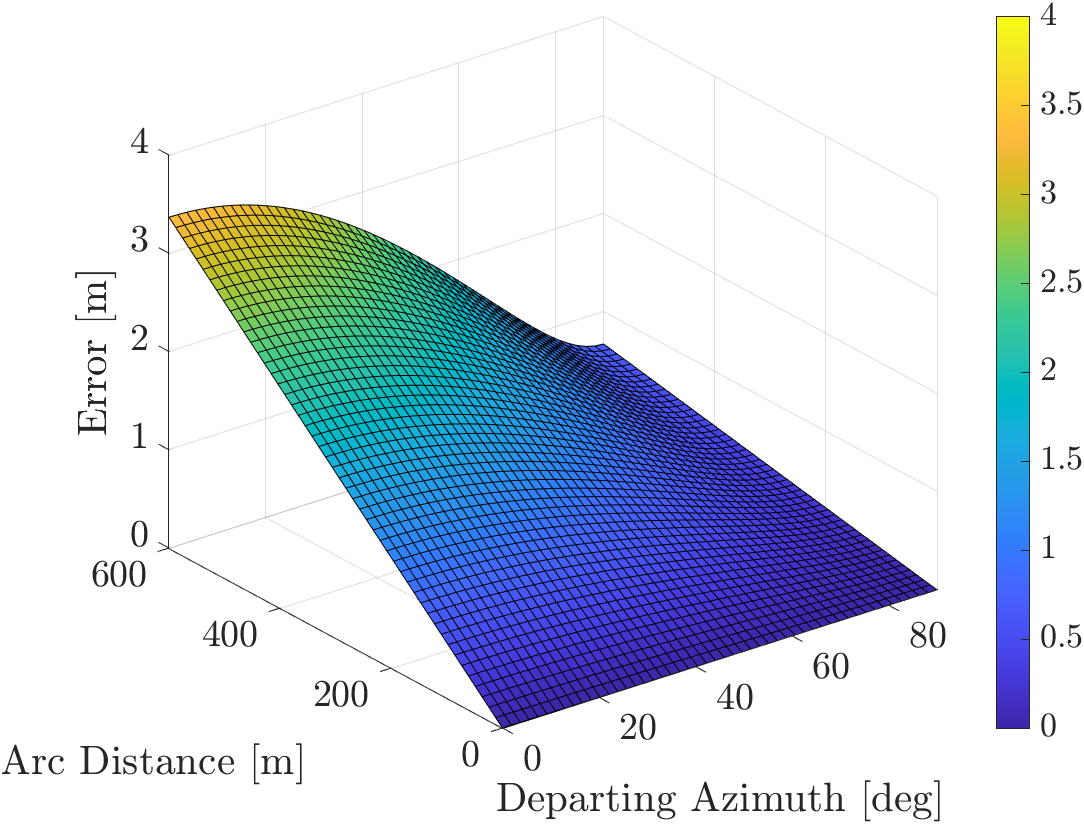}
    \caption{Spherical approximation error as a function of arc length and departure angle.  Error is minimized for zonal trajectories due to the concentric nature of lines of constant latitude for both ellipsoidal and spherical Earth models.}
    \label{fig:spherical_approx_error_departure_angle}
\end{figure}

\newpage 
Over small distances, a spherical approximation provides sufficient accuracy for most real-world marine autonomy applications, while incurring substantially less computational cost than Vincenty's Formulae.  This contention was validated by a 10 million point Monte Carlo simulation.  A uniform distribution of test points was generated upon the surface of the Earth ellipsoid, according to the algorithm described in Appendix \ref{appendix:uniform_sphere_points}.  These points were then matched with a random departure angle and travel distance, permitting the prediction of a new set of geodetic coordinates.  This prediction was carried out using both the spherical position update equations given by Eqs. \eqref{eq:latupdate2} and \eqref{eq:lonupdate3}, and Vincenty's Formulae.  Owning mostly to its iterative nature, Vincenty's solution took more than three times longer to compute than the spherical approximation, on average (when computed on a 2.8 GHz Quad-Core Intel Core i7 processor).

The spherical approximation error -- while generally quite small -- is dominated by initial latitude; mildly influenced by the initial direction of travel; and completely independent of initial longitude, due to Earth's rotational symmetry.  Lines of constant latitude on the WGS84 Earth ellipsoid and the Earth sphere form concentric circles, leading to stronger consistency between the models for zonal (East-West) departure angles.  These results are illustrated in Figs. 5 and \ref{fig:spherical_approx_error_departure_angle}.  For any pair of points on the surface of the earth, the error incurred by the spherical approximation is never greater than 0.56\% the distance traveled, and generally less than 0.41\% ($75^{th}$ percentile).  Thus, for a typical arc length of 10 meters, the error will be less than 5.6 centimeters.

\subsection{Measurement Noise Covariance}
\label{ssec:ukf_R}
Often times in state estimation literature, the process and measurement noise covariance matrices are selected somewhat casually, based mostly on the researcher's intuition, with little explanation or quantitative justification.  Sometimes, estimation of this nature is unavoidable.  However, whenever possible, it is preferable to base these estimates upon real observations of the system of interest.  This reduces the likelihood of over-fitting, and bolsters confidence that the filter design will still meet performance expectations in unforeseen scenarios.

\begin{figure}[ht]
    \centering
    \includegraphics[width=0.6\textwidth]{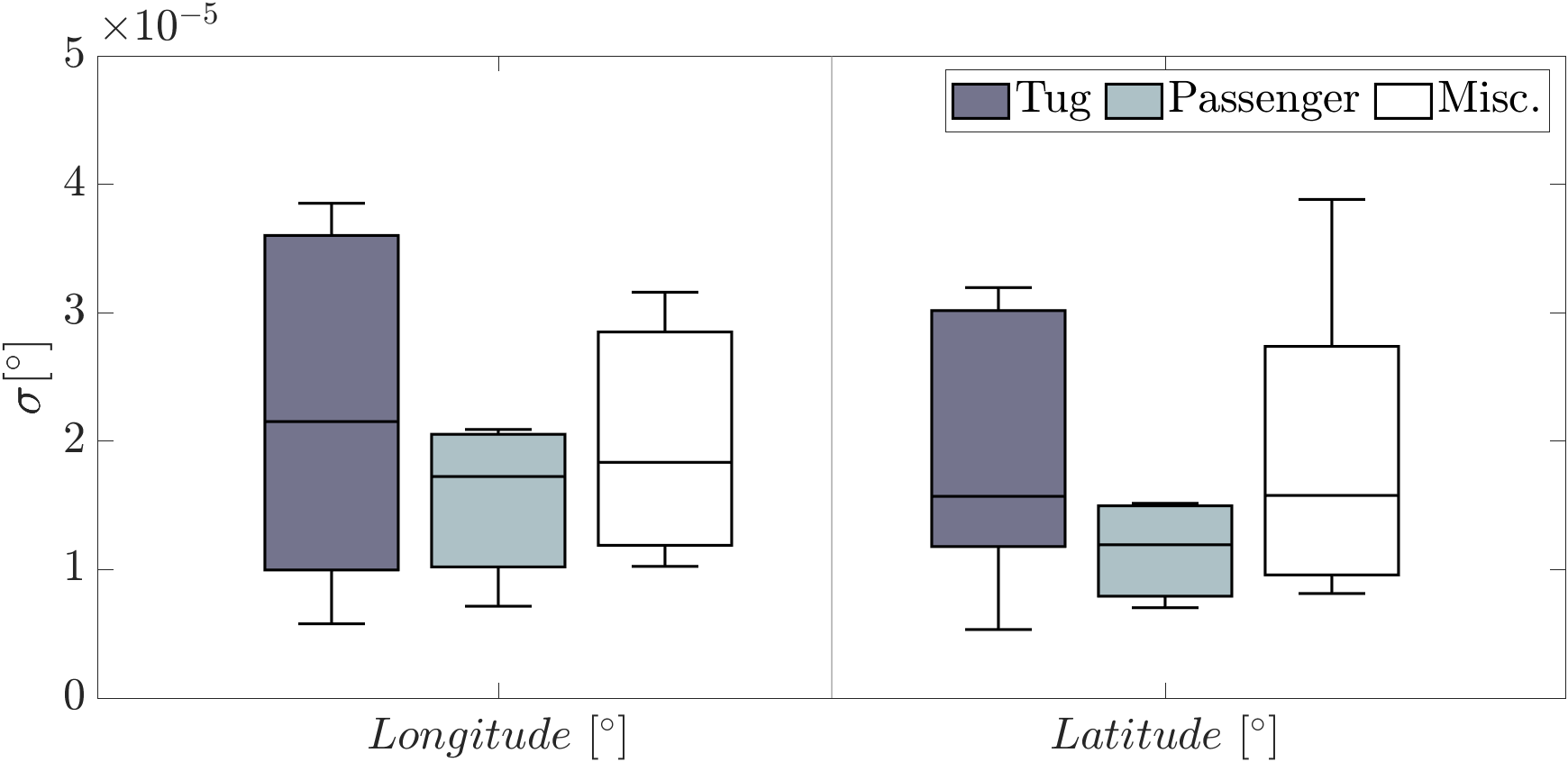}
    \caption{Empirical standard deviation of position for three types of stationary vessels.}
    \label{fig:position-variance}
\end{figure}

To this end, the measurement noise covariance matrix, $\boldsymbol{R}_k$, was justified empirically, using AIS data collected from 21 stationary vessels in Boston Harbor, USA.  Stationary vessels were first identified and separated from moving vessels using a SOG threshold of 0.08 m/s.  This threshold was chosen to accommodate some measurement noise, but the final set of vessels was not particularly sensitive to this value; in other words, it was clear which vessels were docked, and which were underway.  The standard deviation of position was then computed for each stationary vessel.  Finally, each vessel was sorted into one of three categories: \textit{tug}, \textit{passenger}, or \textit{miscellaneous}, to ensure that the precision of AIS position data is consistent across vessel categories.  It was found that position variance is fairly uniform across vessel types, as shown in Fig. \ref{fig:position-variance}.  Accordingly, the position variance was averaged across vessel types, giving standard deviations of $1.90e^{-5}$ degrees longitude ($1.57~m$) and $1.45e^{-5}$ degrees latitude ($1.61~m$) \footnote{Conversion to meters assumes a mean polar radius of $6.371e^{6}~m$, and an equatorial radius of $6.378e^{6}~m$. This is consistent with the WGS84 oblate ellipsoid Earth model.  It is also assumed that Boston Harbor is located at $42.37^{\circ}$ latitude.}.

Unfortunately, it is not possible to estimate the variance of speed and course measurements in this same manner; indeed, one cannot disentangle the influence of process noise from sensor noise while a vessel is underway.  Accordingly, the SOG and COG measurement noise variances were taken from the technical specifications of a commercial dual-antenna GNSS compass.  It is assumed that the noise associated with each AIS field is not correlated; accordingly all off-diagonal elements in the measurement noise covariance matrix are set equal to zero.

\begin{equation}
    \boldsymbol{R} = 
    \begin{bmatrix}
        (1.90e^{-5})^2 & 0      & 0    & 0  \\
        0       & (1.45e^{-5})^2 & 0    & 0  \\
        0       & 0      & (0.05)^2  & 0  \\
        0       & 0      & 0    & (0.2)^2\\
    \end{bmatrix}
\end{equation}

\subsection{Process Noise Covariance}
\label{ssec:ukf_Q}
Estimation of the true process noise covariance, $\boldsymbol{Q}_k$, for all vessel types and dimensions, at all times, is not a trivial undertaking.  One could imagine that a small dinghy, medium sized fishing vessel, and modern post-Panamax shipping vessel would each exhibit drastically different dynamic responses to the same environmental disturbance, due to their vastly different natural frequencies of surge, sway, and yaw.  Furthermore, the severity of these unmodeled disturbances varies both in space and time in the marine environment.  It is possible to estimate the feasible range of wave-induced disturbances using linear wave theory.  Progressive surface-gravity waves produce orbital motion with radius, $\zeta$, and max velocity $|u|$, according to Eqs. \eqref{eq:orbital_rad} and \eqref{eq:orbital_vel}, where $g$ is the gravitational acceleration, $T$ is the wave period, $H$ is the wave height, $k$ is the wavenumber, and $h$ is the water depth respectively \citep{dean1991}.

\begin{gather}
    \zeta = -\frac{H}{2} \frac{\cosh{(k h)}}{\sinh{(k h)}} \label{eq:orbital_rad} \\
    |u| = \frac{g T H k}{4 \pi} \frac{\cosh{(k h)}}{\sinh{(k h)}} \label{eq:orbital_vel}
\end{gather}

In sheltered bays and harbors, this particle excursion is typically very small, and exerts little or no influence on vessel dynamics; however, in the open ocean, wave orbital displacements and velocities can be quite severe.  A few representative sea state scenarios are presented in Table \ref{tab:wave_kinematics}, where the wind fetch was chosen to ensure the evolution of fully-developed seas \citep{bauer1995}.  The peak wave period $T_s$ and significant wave height $H_s$ are derived from the wind speed and wind fetch using the Pierson-Moskowitz Spectrum \citep{pierson1964}, and the water particle kinematics are determined using Eqs.  \eqref{eq:orbital_rad} and \eqref{eq:orbital_vel}.  When the characteristic speed and dimension of the vessel is the same order of magnitude as the speed and radius of wave orbital motion, respectively, the process noise will be substantial.  An ideal tracking solution would leverage an onboard accelerometer to infer the statistical properties of the local wave field, and adjust the process noise covariance matrix accordingly.

At the equator, each degree of longitude spans approximately $111,319.5~m$.  However, as the absolute value of latitude increases, the zonal distance covered by a single degree of longitude decreases according to Eq. \eqref{eq:lon_len}, where $A_0 = 111,319.5~m/^{\circ}$.  Accordingly, the longitude process noise variance must be expressed as a function of vessel latitude.  This ensures that process noise is not underestimated at high-latitudes, where small external disturbances could result in relatively large variations in longitude \footnote{For reference, a two meter zonal arc length is subtended by $1.78e^{-5}$ degrees longitude at the equator, whereas at $70^{\circ}N$, this same arc length is subtended by $5.25e^{-5}$ degrees longitude.}.  The longitude and latitude noise variances are chosen in light of the typical sea states described in Table \ref{tab:wave_kinematics}, in order to compensate for unmodeled wave-induced disturbances of $\zeta_0 = 2m$, at any latitude.

\begin{gather}
    \Delta x_{lon_k} \approx A_0 \cos{(lat_k)} \label{eq:lon_len} \\
    \sigma_{lon_k} = \frac{\zeta_0}{A_0 \cos{(lat_k)}} \label{eq:sigma_lon}
\end{gather}

Unlike the measurement noise, the process noise exhibits correlations between states; specifically, between speed over ground and position (both longitude and latitude).  Furthermore, the sign and magnitude of the correlations depends on the tracked vessel's instantaneous course over ground.  Imagine the following scenario: a vessel is under way, headed due North, when an unexpected southerly wind gust exerts a longitudinal force upon it.  This will result in an unmodeled surge acceleration (assuming a CV model), which will result in an underestimation of both latitude and SOG.  However, if the vessel is on a southbound track, and is nudged southward by a northerly wind, the filter will again underestimate SOG; but, this time it will \textit{overestimate} latitude.  The same phenomena affects longitude estimates, when the vessel is on an eastbound or westbound track.  This coupling is accounted for in the process noise covariance matrix, by allowing the speed-position covariance values to vary with instantaneous COG.

\begin{equation}
\boldsymbol{Q}_k = 
    \begin{bmatrix}
        \sigma_{lon_k}^2 \Delta t_k & 0 & \sigma_{lon_k} \sigma_{\scalebox{.9}{$\scriptscriptstyle U$}} & 0 \\
        0   & \sigma_{lat}^2 \Delta t_k & \sigma_{lat} \sigma_{\scalebox{.9}{$\scriptscriptstyle U$}} & 0 \\
        \sigma_{\scalebox{.9}{$\scriptscriptstyle U$}} \sigma_{lon_k} & \sigma_{\scalebox{.9}{$\scriptscriptstyle U$}} \sigma_{lat} & \sigma_{\scalebox{.9}{$\scriptscriptstyle U$}}^2 & 0 \\
        0 & 0 & 0 & \sigma_{\alpha}^2 \\
    \end{bmatrix} \Delta t_k
\end{equation}
\begin{align*}
    \sigma_{lon_k} &= \frac{\zeta_0}{A_0 \cos{(lat_k)}}\\
    \sigma_{lat} & = \frac{\zeta_0}{A_0}\\
    \sigma_{\scalebox{.9}{$\scriptscriptstyle U$}} & = 0.08 m/s\\
    \sigma_{\alpha} & = 1.2 ^{\circ}\\
    \sigma_{\scalebox{.9}{$\scriptscriptstyle U$}} \sigma_{lon_k} &= \sigma_{lon_k} \sigma_{\scalebox{.9}{$\scriptscriptstyle U$}} =  (\sigma_{lon_k}\sin{(\alpha_k)})^2\\
    \sigma_{\scalebox{.9}{$\scriptscriptstyle U$}} \sigma_{lat} &= \sigma_{lat} \sigma_{\scalebox{.9}{$\scriptscriptstyle U$}} = (\sigma_{lat}\cos{(\alpha_k)})^2\\
\end{align*}

\vspace*{-\baselineskip} 
\begin{table}
\caption{Surface Particle Kinematics for Fully-Developed Sea States (Pierson-Moskowitz Spectrum)}
\label{tab:wave_kinematics}
\centering
\begin{tabular}{lllllll}
\toprule
\textbf{Beaufort Wind Scale} & \textbf{Wind Speed} & \textbf{Wind Fetch} & $\boldsymbol{H_s}$ & $\boldsymbol{T_p}$  & $\boldsymbol{\zeta}$ & $\boldsymbol{|u|}$\\
            & [km/hr]   & [km]   & [m]   & [s]   & [m]   & [m/s]\\
\midrule
\textbf{4}  & 20-28     & 52     & 1.0   & 5.0   & 0.5   & 0.62 \\
\textbf{5}  & 29-38     & 102    & 2.0   & 7.1   & 1.0   & 0.89 \\
\textbf{6}  & 39–49     & 240    & 3.3   & 9.1   & 1.65  & 1.14 \\
\textbf{7}  & 50–61     & 520    & 5.3   & 11.5  & 2.65  & 1.45 \\
\textbf{8}  & 62–74     & 1,110  & 8.2   & 14.3  & 4.10  & 1.80 \\
\textbf{9}  & 75–88     & 1,780  & 11.4  & 16.9  & 5.70  & 2.12 \\
\textbf{10} & 89–102    & 2,980  & 15.5  & 19.7  & 7.75  & 2.47 \\
\bottomrule
\end{tabular}
\end{table}

\subsection{Angular Residual}
\label{ssec:ukf_angular_residual}
The \textit{update} step of a linear Kalman filter requires the computation the arithmetic difference between a state prediction and a contemporaneous observation.  Angular states -- such as heading or COG -- are cyclic on the interval $\big\{\alpha \in \mathbb{R} \mid 0^{\circ} \leq \alpha < 360^{\circ} \big\}$, and thus require special treatment to account for the discontinuity between $359^{\circ}$ and $0^{\circ}$.  For example, if a process model predicts a COG equal to $1^{\circ}$ from true North, but the onboard compass indicates that the vessel COG is $359^{\circ}$, a simple arithmetic difference would indicate a $358^{\circ}$ residual, when it is clear that the true residual error is only $2^{\circ}$.  At best, this will bring about needless degradation of filter confidence, sending erroneous spikes rippling through the covariance matrix; at worst, it could result in filter divergence.  Fortunately, this issue can be rectified by executing Eq. \ref{eq:smallest_signed_error1} after the residual error vector is computed:

\vspace*{-\baselineskip} 
\begin{align}
    \label{eq:smallest_signed_error1}
    \boldsymbol{y}_k &= \boldsymbol{z}_k - \boldsymbol{H}_k \boldsymbol{\Hat{x}}_{k|k-1} \notag \\[3mm]
    y_k^{(\alpha)} &= \big(y_k^{(\alpha)}  + 180 \big) \modd 360 - 180
\end{align}

This method shifts the residual into a continuous, monotonic domain before assessing the angular difference, and then shifts the window back to the desired range.  This method is mathematically rigorous; further, it incurs no adverse effect when the residual does not span the discontinuity at $360^{\circ}$, and thus requires no special logical treatment in such cases.  The output of this algorithm is guaranteed to be in the continuous interval $\big\{y_k^{(\alpha)} \in \mathbb{R} \mid -180 \leq y_k^{(\alpha)} < 180 \big\}$.  This is critically important, as it indicates to the filter in which direction it should adjust its final \textit{a posteriori} estimate: a positive residual indicates the need for a clockwise adjustment, and vise versa.  After the adjusted residual is weighted by the Kalman gain, and fused with the \textit{a priori} state estimate, the COG element of the state vector must be shifted back into the valid range $\alpha \in \left[0:360\right)$, using Eq. \ref{eq:smallest_signed_error2}.  Otherwise, if the model prediction lies east of $360^{\circ}$ by $\Delta\alpha^{\circ}$, while the measurement falls on the westerly side of the discontinuity by a margin greater than or equal to $\Delta\alpha^{\circ} + \delta\alpha^{\circ}$, a negative \textit{a posteriori} COG estimate will result, confounding subsequent recursive COG estimates.

\begin{align}
    \label{eq:smallest_signed_error2}
    \boldsymbol{\Hat{x}}_k &= \boldsymbol{\Hat{x}}_{k|k-1} + \boldsymbol{K}_k \boldsymbol{y}_k \notag\\[3mm]
    \Hat{x}_k^{(\alpha)} &= \Hat{x}_k^{(\alpha)} \modd 360
\end{align}

\section{Simulation}
\label{sec:sim}
In order to verify the efficacy of the proposed tracking filter, a series of computer simulations were performed.  The first simulation, described in Sec. \ref{ssec:sim_typical}, mimics the kinematics of a typical commercial shipping vessel.  The subsequent simulation, described in Sec. \ref{ssec:sim_stability}, is designed specifically to test the practical limits of filter performance and stability.  This simulation is repeated many times, using progressively fewer AIS sensor updates.  The chosen range of simulated AIS update rates cover most typical real-world scenarios, described in Table \ref{tab:AIS_freq} of Appendix \ref{appendix:AIS}.  In each simulation, in order to simulate unmodeled dynamics, Gaussian white noise is added to the SOG and COG states; however, noise is not added to the position states, because doing so produces non-physical kinematic behavior.  Noise is also added to the simulated AIS reports for all states, to mimic measurement noise.  The measurement noise is modeled based on typical commercial GNSS specifications.

The performance of the proposed geodetic UKF is compared with that of a plane-Cartesian EKF.  To this end, a local tangent plane is defined, with origin at $[lon_0 = -71.0237 \mathrm{^{\circ}E}, ~lat_0 = 42.3469 \mathrm{^{\circ}N}]$.  The basic EKF structure is borrowed from \cite{fossen2018a}, who describe a plane-Cartesian EKF with the dynamics, state vector, process model and Jacobian matrix defined by Eqs. \eqref{eq:ekf_dynamics}, \eqref{eq:ekf_observer}, \eqref{eq:ekf_state}, \eqref{eq:ekf_model}, and \eqref{eq:ekf_jacobian}.  Both filters are run at a rate of one iteration per second (1 Hz), and are given access to simulated noisy AIS data every 6 seconds (0.17 Hz).

\begin{align}
    \boldsymbol{\Dot{x}} &= \boldsymbol{f}(\boldsymbol{x}) + \boldsymbol{Bu} + \boldsymbol{w} \label{eq:ekf_dynamics}\\
    \boldsymbol{y} &= \boldsymbol{Cx} + \boldsymbol{e} \label{eq:ekf_observer}
\end{align}
\begin{gather}
    \boldsymbol{x} = [x, y, U, \mathcal{X}] \label{eq:ekf_state} \\
    \boldsymbol{f}(\boldsymbol{x}) =
    \begin{bmatrix}
        x_3 \cos{(x_4)} \\  x_3 \sin{(x_4)} \\ 0 \\ 0
    \end{bmatrix} \label{eq:ekf_model} \\
    \boldsymbol{A} = \dfrac{d \boldsymbol{f}(\boldsymbol{x})}{d \boldsymbol{x}} = 
    \begin{bmatrix}
        0       & 0     & \cos{(x_4)}   & -x_3 \sin{(x_4)}  \\
        0       & 0     & \sin{(x_4)}   & x_3 \cos{(x_4)}  \\
        0       & 0     & 0             & 0  \\
        0       & 0     & 0             & 0\\
    \end{bmatrix} \label{eq:ekf_jacobian}
\end{gather}

The vector-valued variables $\boldsymbol{w}$ and $\boldsymbol{e}$ are the Gaussian process and measurement noise, respectively; the measurement matrix, $\boldsymbol{C} = \boldsymbol{I}$; the initial covariance, $\boldsymbol{P}_0 = 0.1\boldsymbol{I}$; process noise covariance matrix, $\boldsymbol{Q} = diag([0.01; 0.01; 0.1; 0.1])$; and measurement noise covariance matrix, $\boldsymbol{R} = diag([1e^{-3}; 1e^{-3}; 1e^{-3}; 1e^{-2}])$.  If the position process noise variance values strike the reader as large, recall that they are described in a local Cartesian coordinate frame, and thus have units of meters-squared rather than degrees-squared.  This EKF approach recommends using a constant-velocity (CV) motion model when the sensor update rate is slower than 0.5 Hz; accordingly, acceleration and turn-rate are set equal to zero for both the EKF ($\boldsymbol{u} = 0$), and geodetic UKF ($a_k = r_k = 0$), for all simulations.  The goal of this comparison is not to prove the superiority of the geodetic UKF approach, but rather to show \textit{comparable} performance under typical operating conditions.  Indeed, if performance parity can be demonstrated, it is argued that the additional features of the geodetic UKF -- enhanced ease of use, interpretability, stability, and computational efficiency -- make it better suited for use on long-distance crewed and autonomous platforms.

\subsection{Typical Maneuver}
\label{ssec:sim_typical}
\begin{figure}[ht]
    \centering
    \includegraphics[width=0.6\textwidth]{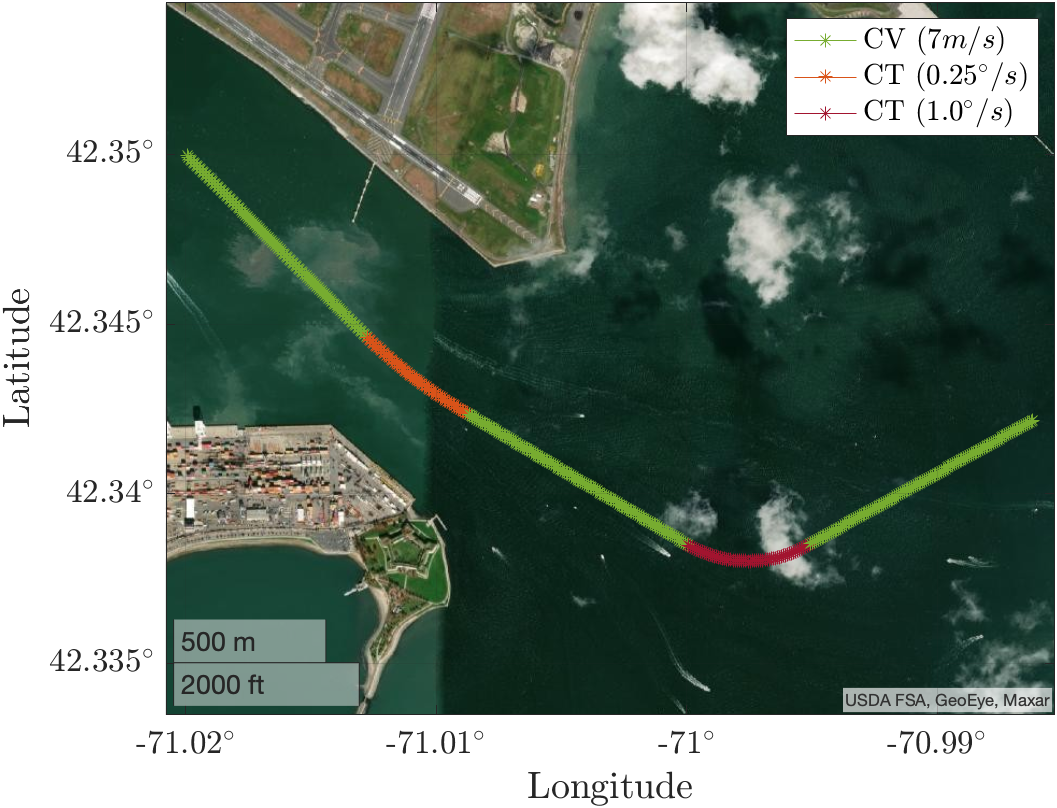}
    \caption{Simulated vessel trajectory, departing Boston Harbor from West to East.}
    \label{fig:bos_sim_path}
\end{figure}

The first simulation mimics the trajectory of a real cargo vessel, Glovis Chorus (MMSI\# 440292000), which departed Boston Harbor, USA, on June 8th, 2020.  The simulated vessel path is shown in Fig. \ref{fig:bos_sim_path}.  The simulated vessel moves through a series of constant velocity (CV) and constant-turn (CT) maneuvers, at a constant forward speed of $7 \mathrm{~m/s}$.  In order to verify that the geodetic UKF was functioning properly, the absolute position error was overlayed upon the $3\sigma$ bounds inferred from the trace of the covariance matrix.  Fig. \ref{fig:bos_sim_pos_error1} shows that the position estimation error always lies well within the expected theoretical bounds, confirming that the process model is able to capture the vessel kinematics.

\begin{figure}[ht]
    \centering
    \includegraphics[width=0.55\textwidth]{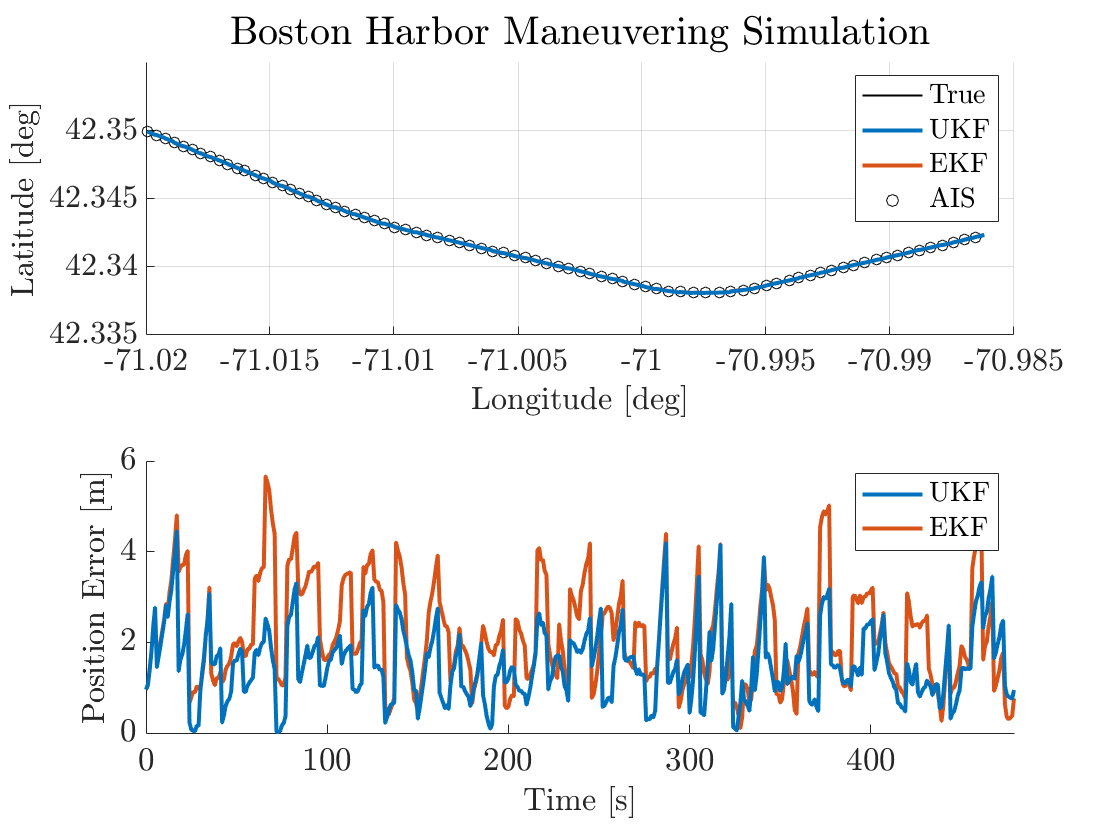}
    \caption{A comparison of absolute position error for the geodetic UKF and planar EKF tracking filters, with $\Delta t_{AIS} = 6 \mathrm{~s}$..  In the upper figure, the true vessel trajectory and the trajectories estimated by both the EKF and UKF are nearly identical; accordingly, only the UKF trajectory is visible.}
    \label{fig:bos_sim_pos_error2}
\end{figure}

\begin{figure}[ht]
    \centering
    \includegraphics[width=1\textwidth]{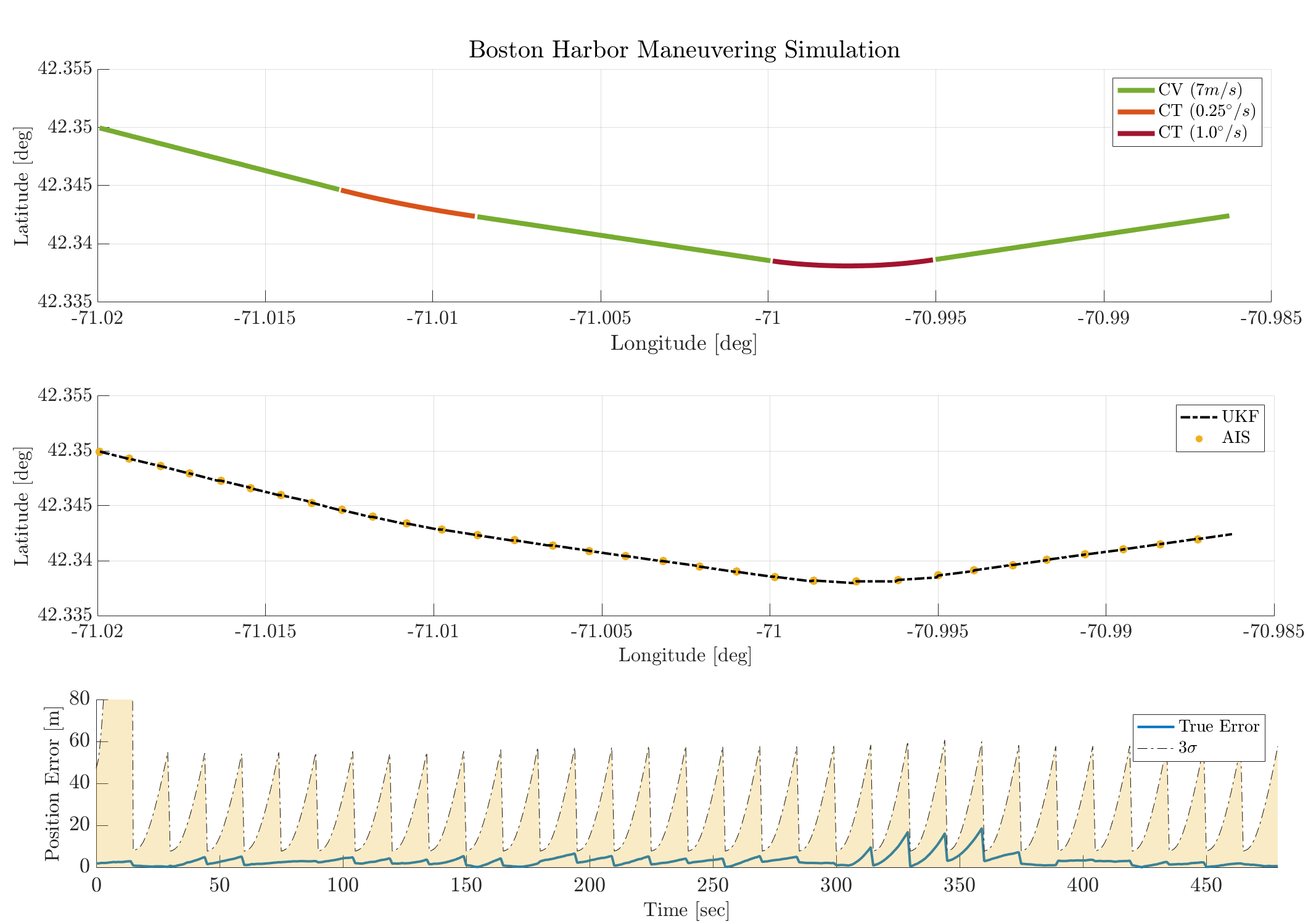}
    \caption{Absolute position error for the geodetic UKF using a $1 \mathrm{~Hz}$ update frequency, with $\Delta t_{AIS} = 15 \mathrm{~s}$.  The simulated vessel speed is $7 \pm 0.1 \mathrm{~m/s}$. The distance between the estimated position and true position is determined by solving Vincenty's Inverse Problem \citep{vincenty1975}.  The $3\sigma$ bounds are inferred from the trace of the covariance matrix, and show that the state estimation error is always well within the expected theoretical limit.}
    \label{fig:bos_sim_pos_error1}
\end{figure}

Fig. \ref{fig:bos_sim_pos_error2} and Table \ref{tab:bos_sim_error} indicate that the tracking performance of the proposed geodetic UKF is comparable to that of a typical plane-Cartesian EKF, for all states.  However, recall that the EKF requires many geodetic transformations to achieve this result.  Each time an AIS message is received, its geodetic position report must first be expressed in Earth-Centered Earth-Fixed (ECEF) coordinates, and then mapped onto a local North-East-Down (NED) tangent plane.  In order to validate EKF vessel position estimates, they must be converted back into geodetic coordinates.  As the number of tracked vessels grows, the additional computational burden of these frequent geodetic conversions can become substantial.  Operating in a busy harbor may require an ASV to track dozens of nearby vessels, and projecting each vessel's geodetic position onto a planar coordinate frame may incur thousands of unnecessary trigonometric function calls per second. The proposed geodetic UKF describes the vessel kinematics directly in geodetic coordinates, and does not require any such transformations.

\newpage
\begin{table}
\caption{RMS Error for Boston Harbor Tracking Simulation: Planar EKF vs. Geodetic UKF Comparison}
\label{tab:bos_sim_error}
\centering
\begin{tabular}{lllll}
\toprule
        & \textbf{LON}  & \textbf{LAT}  & \textbf{SOG}  & \textbf{COG} \\
        & [deg]         & [deg]         & [m/s]         & [deg]\\
\midrule
\textbf{UKF}    & $1.25e^{-5}$   &  $1.24e^{-5}$    &   $0.13$     &    $2.031$ \\
\textbf{EKF}    & $1.87e^{-5}$   &  $1.67e^{-5}$    &   $0.14$     &    $2.096$ \\
$\Delta$        & $-0.62e^{-5}$  & $-0.43e^{-5}$    &  $-0.01$   &   $-0.065$ \\
\bottomrule
\end{tabular}
\end{table}

\null\vfill
\vfill\null
\newpage

\subsection{Stability Analysis}
\label{ssec:sim_stability}
The ``lawnmower pattern'' is a typical trajectory followed by vessels conducting bathymetric surveys, hydrographic research, and mine-sweeping exercises.  However, because it involves periodic sharp turns, it also a useful pattern for testing the stability of a tracking filter.  For systems with very frequent sensor updates, these sudden maneuvers pose little or no challenge; indeed, over the span of one measurement, the trajectory is mostly static.  However, for systems with access to only intermittent, sparse measurements, a sudden unexpected maneuver can cause filter divergence.

\begin{figure}[ht]
    \centering
    \includegraphics[width=0.58\textwidth]{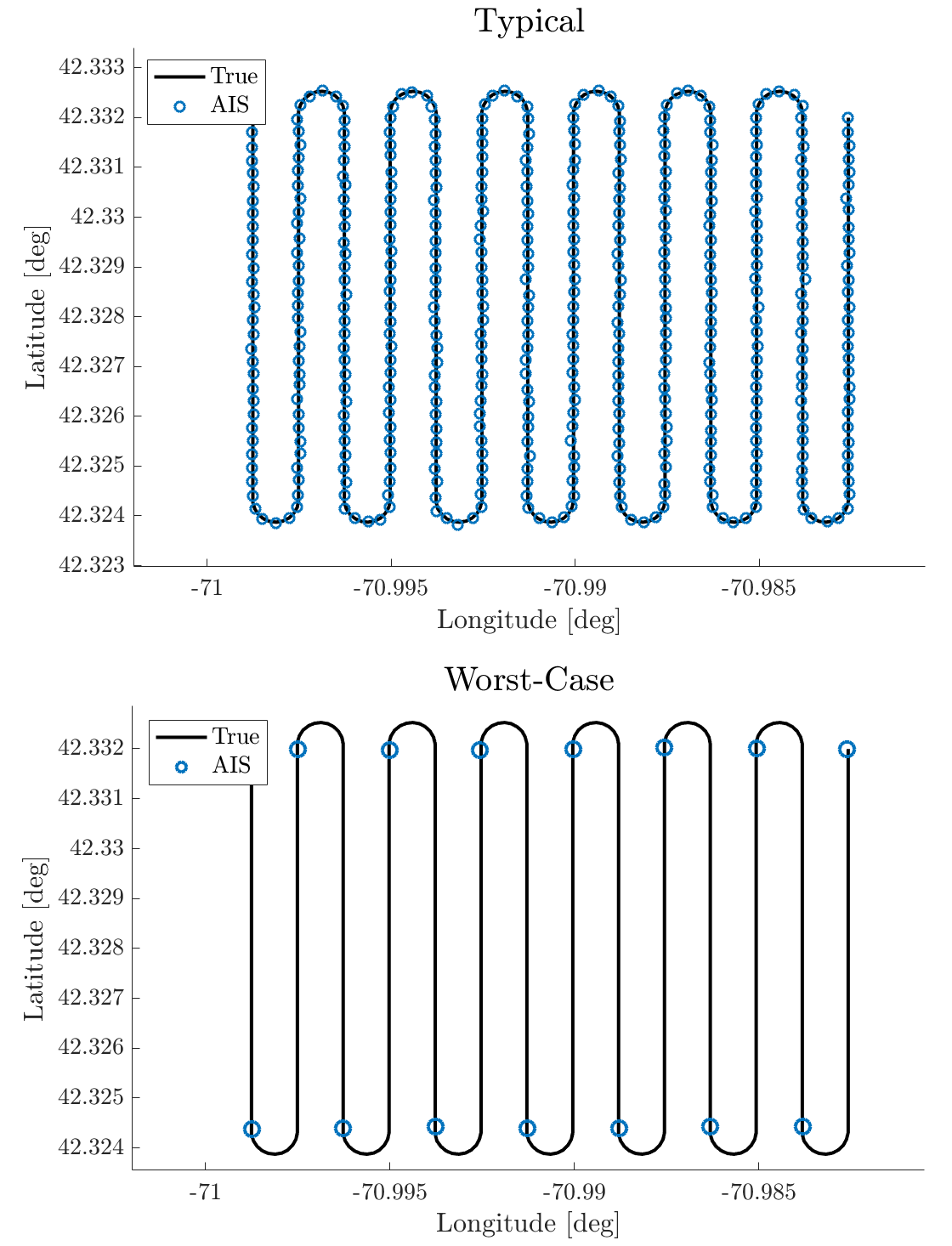}
    \caption{Simulated ``lawnmower'' vessel trajectory, designed to test the stability of the proposed filter.  The ``worst case scenario'' corresponds to a situation in which the AIS transmission interval exactly matches the temporal periodicity of the lawnmower maneuvers.}
    \label{fig:lawnmower}
\end{figure}

In order to test the stability of the proposed filter, a lawnmower pattern simulation was conducted many times.  The specific radius of the turning maneuvers is not particularly important, but it should be small enough (relative to vehicle speed and the length of the straightaway sections) to illustrate the deleterious effect of a sudden maneuver on filter performance.  For these simulations, the straightaway length is $855 \mathrm{~m}$, and the turning radius is $50 \mathrm{~m}$.  The ratio of vessel speed to the length of each repeated section (including one straightaway, and one 180 degree turn) is also important, as it is a proxy for the severity of the maneuvers.  The simulated vessel speed was held constant at $15 \pm 0.1 \mathrm{~m/s}$, across all simulations.  The number of repeated sections was varied between ten and thirty, and did not substantially effect the state estimation root mean squared error (RMSE).  However, it is important to increase the number of repeated maneuvers until the RMSE for each state becomes mostly static; otherwise, one risks missing a filter divergence event resulting from the accumulated error of multiple maneuvers. 

\begin{figure}[ht]
    \centering
    \includegraphics[width=0.65\textwidth]{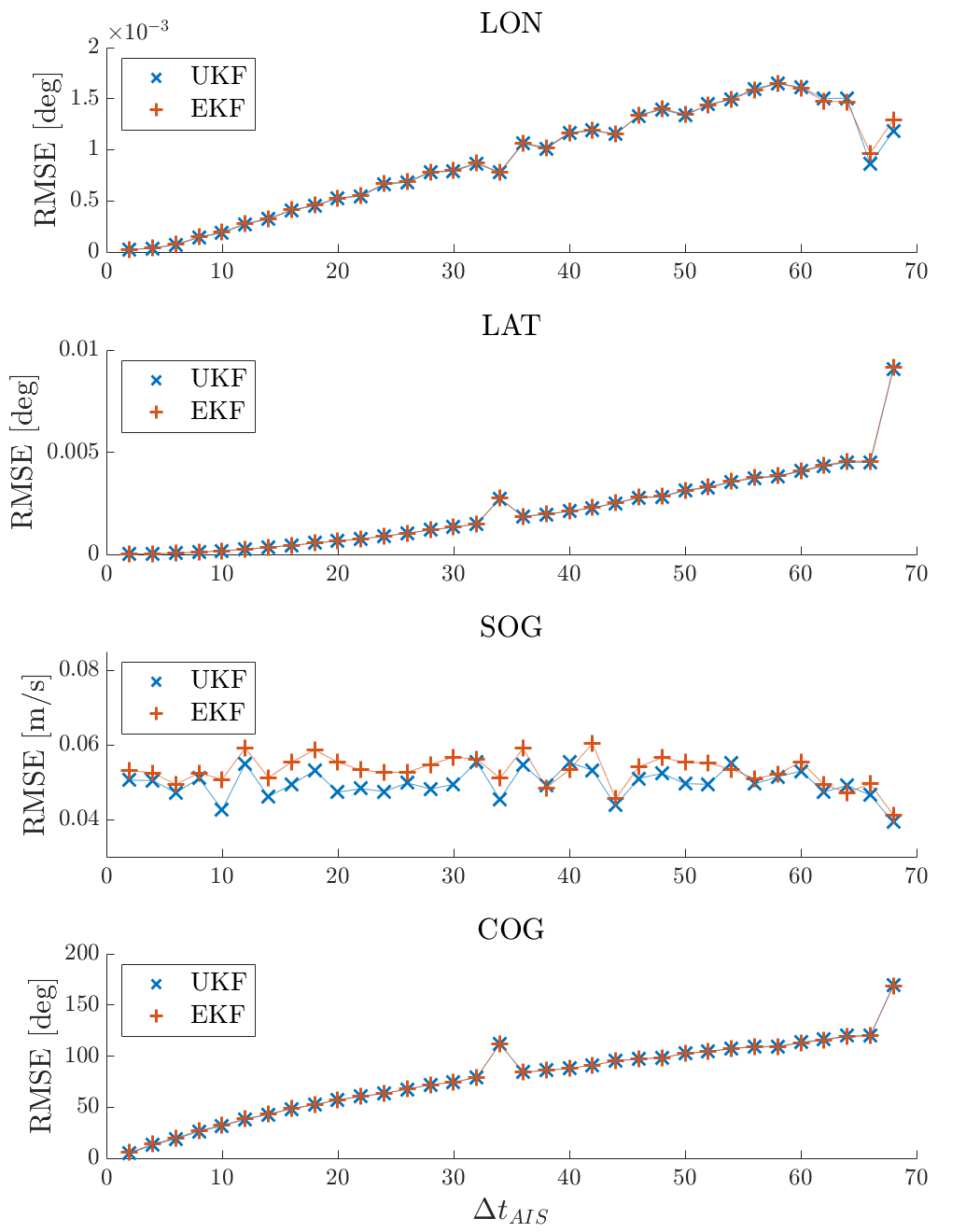}
    \caption{Absolute state error for the geodetic UKF and planar EKF tracking filters, using a $1 \mathrm{~Hz}$ update frequency.  In order to test stability, the simulated lawnmower pattern is repeated many times, with progressively fewer AIS updates.  The simulated vessel speed is $15 \pm 0.1 \mathrm{~m/s}$.  Both the EKF and UKF remain stable for all AIS update rates; any further increase in the AIS transmission interval beyond $\Delta t_{AIS} = 68 s$ incurs less error due to spatial aliasing.}
    \label{fig:lawnmower_error}
\end{figure}

For each successive simulation, the AIS sensor update period was increased by one second, from an initial two seconds, to a final sixty-eight seconds.  This final value was chosen specifically to match the exact temporal period of the lawnmower pattern, resulting in a ``worst-case scenario,'' in which the vessel transmits an AIS measurement only moments prior to executing a rapid 180 degree turn.  This results in the maximum possible residual error in the COG estimate, as well as large residual errors in position that are proportional to the length of the straight sections.  The SOG residual error is unaffected by the AIS update frequency, as it its held constant throughout the entire simulation.  The true vessel position, and the location of each AIS update, is shown in Fig. \ref{fig:lawnmower}, for both the best case scenario (upper figure, $\Delta t_{AIS} = 2 s$) and worst case (lower figure, $\Delta t_{AIS} = 68 s$) simulations.

Not surprisingly, as the AIS update frequency decreases, the RMSE increases.  This intuitive relationship is illustrated in Fig. \ref{fig:lawnmower_error}.  There are two important aspects contained in this result: first, the proposed geodetic UKF matches the performance of a traditional plane-Cartesian EKF; second, neither filter becomes unstable, even when forced to contend with the highly improbable worst case scenario.  Indeed, for the most part, error increases smoothly and linearly with increasing AIS update period.  Note that in this specific scenario, the RMSE does not continue to grow for AIS update intervals in excess of $\Delta t_{AIS} = 68 s$; indeed, beyond this worst case scenario, the filter begins to alias the lawnmower pattern, resulting in a lower apparent RMSE.

This series of simulations are intended to probe filter performance under a truly unlikely scenario: one in which a tracked vessel is both maneuvering frequently, and transmitting AIS updates infrequently.  In practice, most commercial vessels transmit AIS updates at an interval much closer to six seconds than sixty.  However, it is nevertheless a useful exercise to assess the performance and stability of a new tracking filter under a worst-case scenario.  

\section{Field Test}
\label{sec:field}

\begin{figure}[ht]
    \centering
    \includegraphics[width=0.5\textwidth]{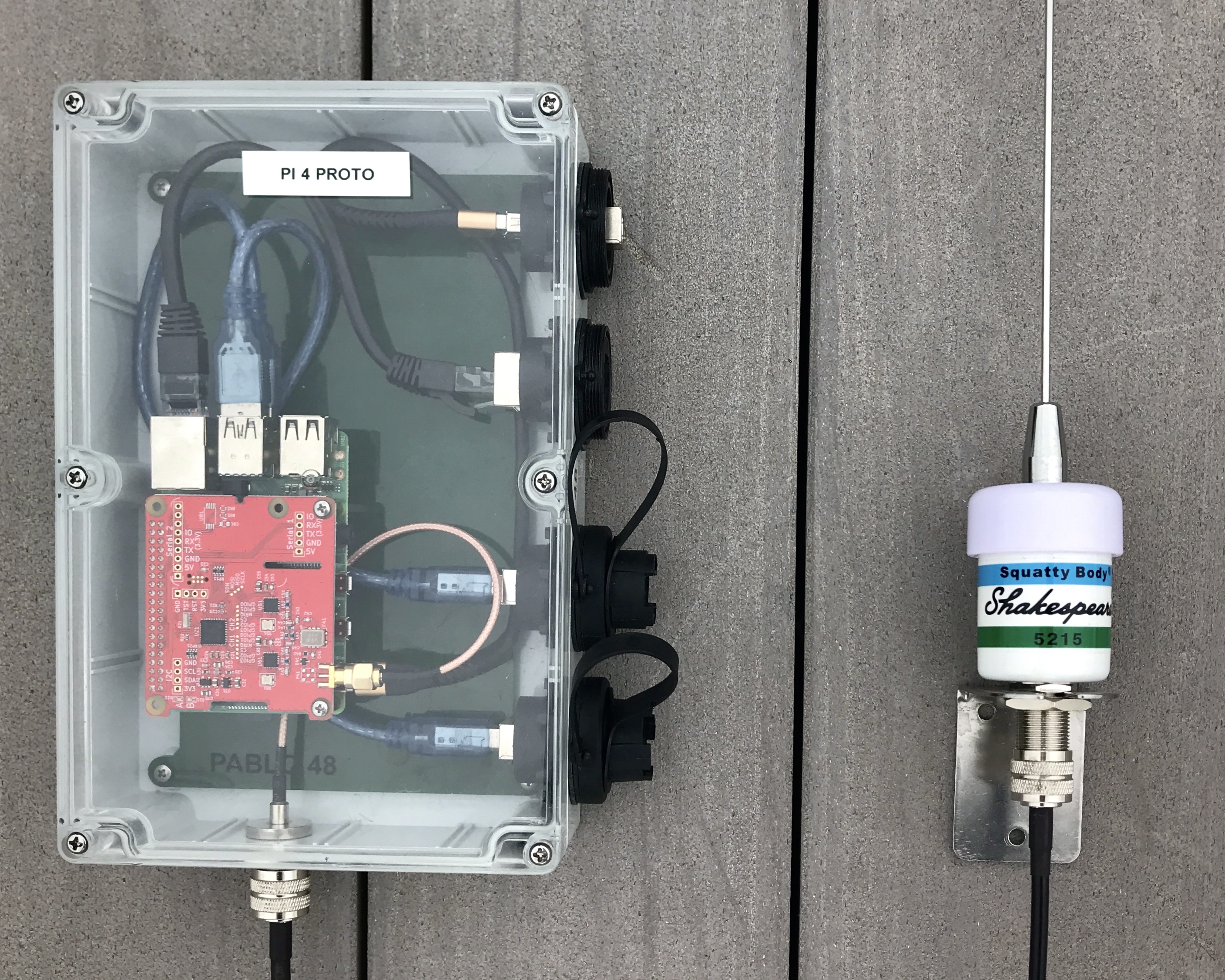}
    \caption{The data collection apparatus consists of an AIS receiver and Raspberry Pi 4B computer mounted in a waterproof enclosure, connected to a 36" VHF antenna via a low-loss coaxial cable.}
    \label{fig:exp_setup}
\end{figure}

AIS data was collected at a position near the mouth of Boston Harbor, USA, using a Raspberry Pi Model 4B, outfitted with a \textit{dAISy HAT} AIS Receiver, manufactured by Wegmatt, LLC.  The \textit{dAISy HAT} is a low-power, two-channel A/B (161.975 MHz, 162.025 MHz) AIS receiver, which communicates with the Raspberry Pi via UART, using the exposed GPIO pins, as shown in Fig \ref{fig:exp_setup}.  The receiver is also capable of communicating with any generic computer via USB.  The AIS receiver is also connected to a \textit{Shakespeare 5215} 36" VHF antenna via low-loss coaxial cable (PL-259), and a pair of SO-239 connectors.  The receiver and computer are both powered by a GoalZero \textit{Flip20} 5200 mAh lithium-ion battery.  Under typical operating conditions, this small battery is capable of powering the entire system for 7.2 hours.  The lion's share of this energy expenditure can be attributed to the background processes running on the Raspberry Pi computer; indeed, the AIS receiver itself draws less than 40 mA at 5V when operating in receive mode.  This suggests an AIS receiver can be implemented on essentially any existing ASV, and incur an almost negligible impact on the platform's total energy budget.

A number of pre-processing steps were required to convert the AIS position reports into a interpretable form.  These steps are outlined for convenience in Appendix \ref{appendix:AIS}.  Fig. \ref{fig:field_ais} shows the trajectories of five vessels underway in Boston Harbor on June 8th, 2020.  The AIS data collected for these vessels was used to test the real-world performance proposed geodetic UKF, and contrast its performance with that of a comparable plane-Cartesian EKF.  AIS data is received at irregular intervals, due to VHF transmission losses, corruption due to multi-path errors, and occlusion effects; accordingly, the proposed filter was designed to process data asynchronously, repeating its prediction step until a new AIS report was available.

The proposed geodetic UKF performed admirably.  While true error cannot be determined in field experiments, the residual error was in line with expectations ($<3\sigma$).  Table \ref{tab:field_error} indicates that the UKF and EKF achieved nearly identical performance on the field data.  It is important to recall, however, that if these vessels were to travel away from the local origin, the accuracy of the EKF solution would begin to deteriorate, while the performance of the geodetic UKF would remain consistent.  In Table \ref{tab:field_error}, the large position errors shown in rows three and five were the result of sparse AIS updates.  The performance of any AIS-based tracking filter depends directly upon the frequency and accuracy of AIS reports.  The proposed tracking filter is not intended to be a comprehensive solution for vessel detection and tracking; but rather, an inexpensive starting point that leverages the AIS protocol to provide crucial situational awareness to otherwise blind ASVs.

\newpage
\begin{table}
\caption{Residual RMSE of Position Estimates}
\label{tab:field_error}
\centering
\begin{tabular}{lllll}
\toprule
\textbf{MMSI}   & \textbf{VESSEL NAME}  & \textbf{UKF}  & \textbf{EKF}  & $\Delta$\\
                &                       & [m]           & [m]           & [m]\\
\midrule
\textbf{367513010}    & JUSTICE                 &  10.28    &   10.26   &    0.02 \\
\textbf{367029470}    & BOSTON PILOT CHELSEA    &  7.39     &   7.45    &    -0.06 \\
\textbf{338046826}    & SEATOW RESCUE 2         &  328.86   &   328.24  &    0.62 \\
\textbf{440292000}    & GLOVIS CHORUS           &  4.17     &   4.22    &    -0.05 \\
\textbf{538070509}    & BOLERO                  &  155.2    &   155.4   &    -0.2 \\
\bottomrule
\end{tabular}
\end{table}

\begin{figure}[ht]
\label{fig:field_ais}
\begin{center}
    \centering
    \includegraphics[width=0.94\textwidth]{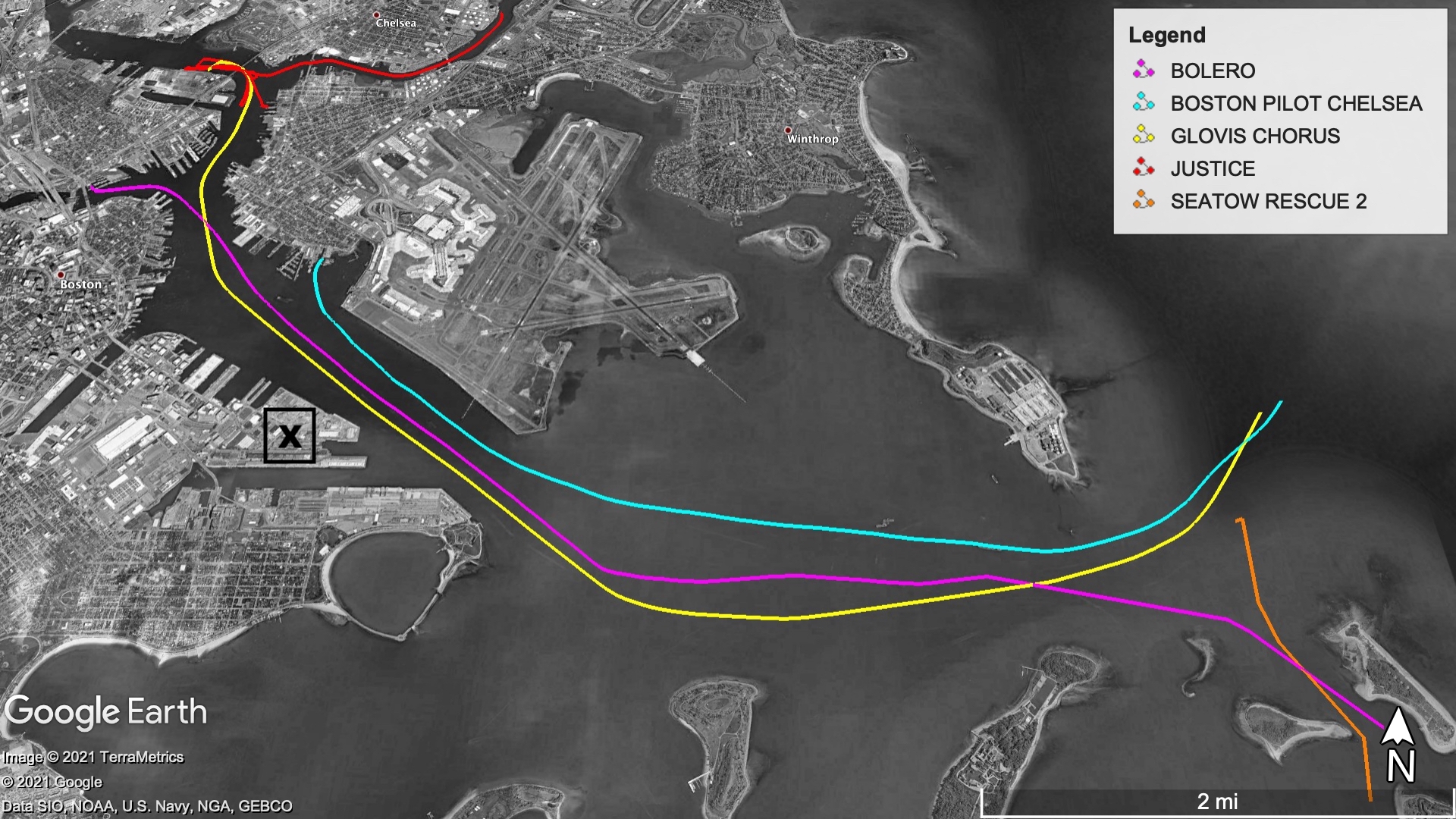}
    \caption{AIS data collected near the entrance of Boston Harbor.  The AIS receiver location is marked by a boxed X.}
\end{center}
\end{figure}

\section{Conclusion}
This work describes a novel means of maritime vessel tracking, suitable for use on a wide variety of both crewed and autonomous surface vehicles.  It is best characterized by its ability to track targets directly in geodetic coordinates, eliminating the need for local planar coordinate systems.  This simplifies the tracking problem, reduces computational overhead, and eliminates linearization error.  The proposed tracking filter is an inexpensive, low-power, plug-and-play solution, and is thus particularly well suited to small and medium-sized ASVs.  It is important to emphasize that the goal of this work is not to prove that the geodetic UKF is superior to a plane-Cartesian EKF under all circumstances, nor that the use of a local coordinate system is always inappropriate; rather, the authors hope to convey that there are concrete advantages to using the geodetic UKF, especially in the context of long-distance surface vessel tracking.  First, due to the inherently nonlinear kinematics of vessel motion on an ellipsoidal shell (and the fact that AIS data is reported in geodetic coordinates) some degree of model nonlinearity is unavoidable: it must be accounted for in either the process model, or the measurement model.  It is argued that it is both more convenient, and more computationally efficient, to absorb this nonlinearity in a geodetic process model.  This filter design choice enables the use of a simple linear measurement model, and eliminates the need for any pre-processing steps (i.e. mapping geodetic AIS data onto a local coordinate frame).  Second, describing the target kinematics directly in geodetic coordinates eliminates the linearization error incurred by the projection of geodetic positions onto a local planar reference frame.  While this source of error is small near the origin, it grows quickly with increasing vessel separation and distance from the origin (see Appendix \ref{appendix:plane_error}).  Prior-EKF based approaches must contend with this error by periodically redefining the local coordinate frame (either manually or algorithmically), which incurs some computational cost, and also makes the operating region less interpretable.

The usefulness of the described methodology extends beyond its applicability for small unmanned surface platforms; indeed, the recent preponderance of high-profile military and cargo vessel collisions suggests that crewed vessels, too, could benefit from enhanced AIS-based vessel tracking and trajectory prediction \citep{miller2019}.  AIS is less subject to occlusion effects than RADAR and LIDAR, due to the robust propagation characteristics of VHF radio waves.  While line-of-sight sensors play an invaluable role in marine vessel detection and tracking, they are expensive, and often require human experts to operate correctly.  The proposed methodology can serve as either a low-cost alternative to line-of-sight sensors, or a complementary feature to be used in conjunction with such sensors.  The latter is the preferred approach, and could provide comprehensive situational awareness for both crewed and autonomous surface vessels.

\section*{Acknowledgment}
The authors would like to thank Dr. Supun Randeni and Dr. Casey Handmer for their technical insights; Dr. Michael R. Benjamin for provisioning the authors with hardware and data acquisition equipment; and Dr. Henrik Schmidt for his continued mentorship and guidance.

\clearpage
\appendix
\section{Linearization Error in Local Planar Coordinates}
\label{appendix:plane_error}
A first order approximation of the error incurred by projecting geodetic data onto a planar coordinate system can be attained using spherical geometry.  The projected distance from the origin, $L$, as a function of true distance from the origin, $s$, is given by Eq. \ref{eq:plane_proj}, where $\theta$ is the angular separation between the point and the origin in spherical coordinates, and $R$ is the radius of the sphere.  Note that only one angular coordinate is required due to spherical symmetry.

\begin{equation}
    \label{eq:plane_proj}
    L = R \sin{(\theta)} = R \sin{\left( \frac{s}{R} \right)}
\end{equation}

\begin{figure}[ht]
    \centering
    \includegraphics[width=0.4\textwidth]{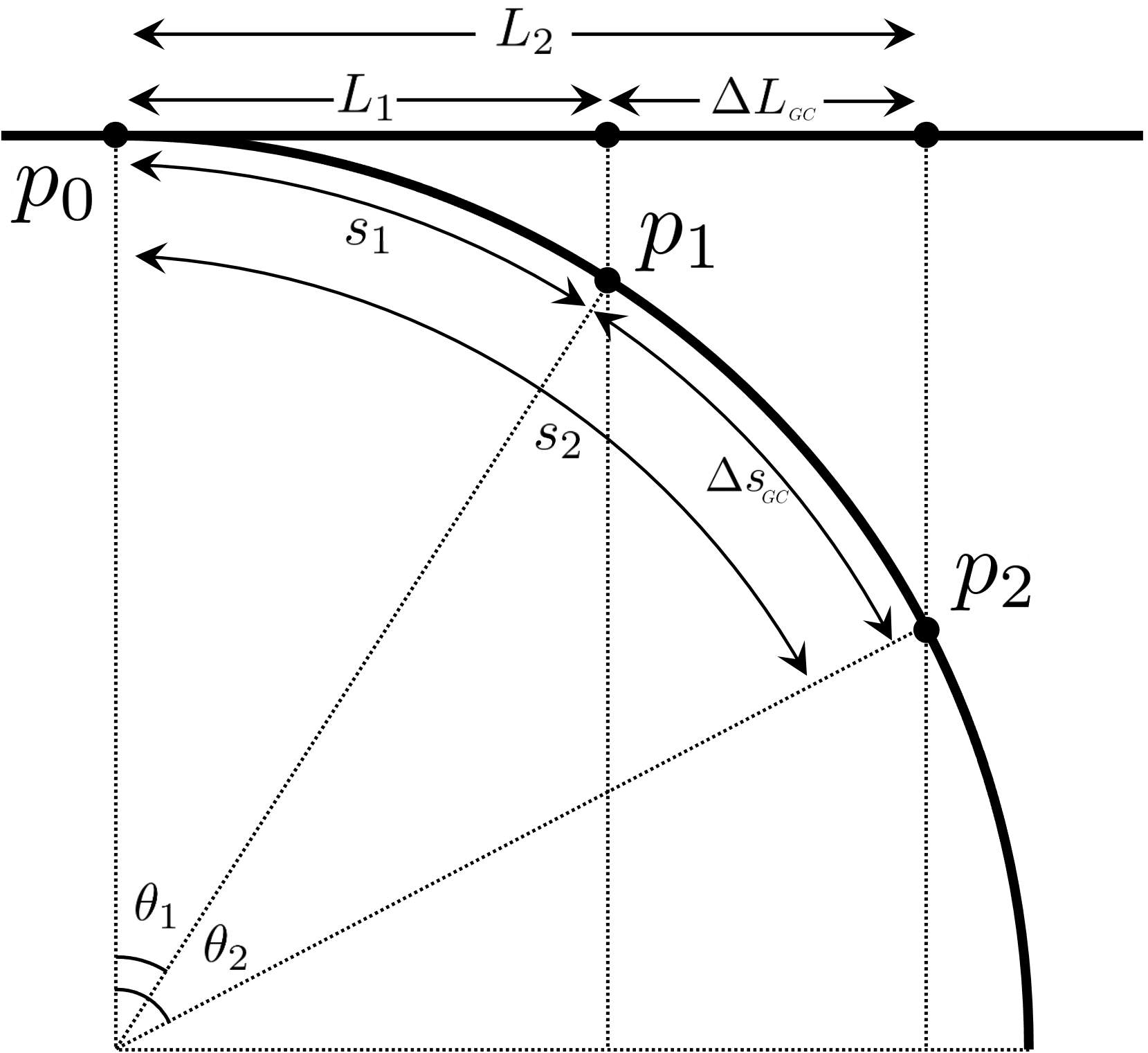}
    \caption{Two points upon a shared great circle path.  The separation error, $\epsilon_{_{GC}}$, is a function of both the absolute separation, $s_2-s_1$, and the mean distance of the points from the origin, $\frac{s_1+s_2}{2}$.  Error grows slowly at first, and then rapidly with increasing point separation and mean distance from the origin.}
    \label{fig:tangent_plane_GC}
\end{figure}

Thus, absolute position error is attained by subtracting $L$ from $s$.  However, for many guidance and navigation applications (including collision avoidance), the absolute position error of a single point is not as important as the range error (also referred to as ``separation error'') between points.  In the simplest case, take two points, $p_1$ and $p_2$, which lie upon the same great circle, each some distance from the origin of the local planar coordinate system ($p_0$), as shown in Fig. \ref{fig:tangent_plane_GC}.  A so-called ``great circle path'' is defined by the intersection of a sphere and any plane coincident with the center of the sphere, and thus forms a circle in $\mathbb{R}^3$.  For such a scenario, the great circle separation error, given by Eq. \ref{eq:plane_slice_error}, is simply the difference between the true separation (given by the difference of two arc lengths, $s_1$ and $s_2$) and the projected separation (given by the difference of two line segments, $L_1$ and $L_2$).  These two quantities are computed using Eqs. \ref{eq:plane_slice_dL} and \ref{eq:plane_slice_ds}.

\begin{align} 
    \Delta L_{_{GC}} &= |L_2 - L_1| \\
    &= R \left| \sin{\left( \frac{s_2}{R} \right)} - \sin{\left( \frac{s_1}{R} \right)} \right| \\
    &= 2R \left| \sin{\left( \frac{s_2-s_1}{2R} \right)} \cos{\left( \frac{s_1+s_2}{2R} \right)} \right| \label{eq:plane_slice_dL}\\
    \Delta s_{_{GC}} &= |s_2 - s_1| \label{eq:plane_slice_ds}\\
    \epsilon_{_{GC}} &= \Delta s_{_{GC}} - \Delta L_{_{GC}} \label{eq:plane_slice_error}
\end{align}

The absolute value brackets were removed from Eq. \ref{eq:plane_slice_error}, due to the fact that $\Delta s_{_{GC}} > \Delta L_{_{GC}}$ for all pairs of non-coincident points on a great circle.  These equations reveal that linearization error is negligible for pairs of points that lie both close to the origin, and to each other.  However, error begins to grow very quickly if either of these parameters -- point separation or mean distance from the origin -- become too large.

\begin{figure}[ht]
    \centering
    \includegraphics[width=0.5\textwidth]{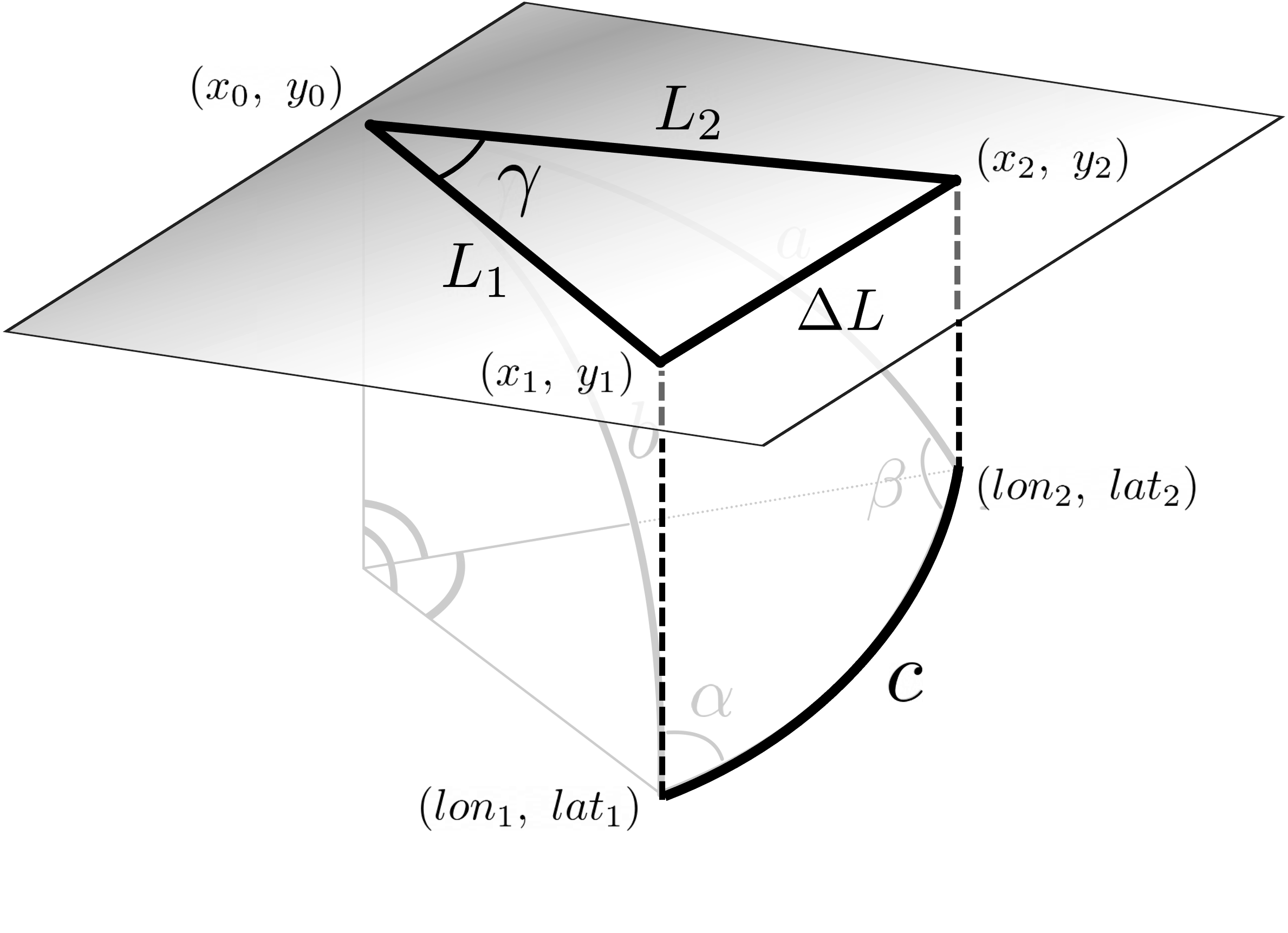}
    \caption{A modified version of the spherical triangle shown in Fig. \ref{fig:sph_triangle}, emphasizing the effect of linear approximations on geodetic positions and distances.  In the planar reference frame, position is defined in the traditional Cartesian manner, using two orthonormal basis vectors.  As each vessel strays farther from the origin, position error increases in a nonlinear fashion.  This is particularly concerning in collision avoidance scenarios, as there may be a discrepancy between the true vessel separation (given by the arc length $c$), and the approximate separation (given by the linear segment $\Delta L$).  The magnitude of this error depends strongly upon the vessels' angular separation, $\gamma$, and their projected distances from the origin, $L_1$ and $L_2$.}
    \label{fig:sph_triangle_tangent_plane}
\end{figure}

While this ``two-plane'' scenario is useful for building intuition, it does not capture the full error dynamics of the nonlinear sphere-to-plane mapping.  Rather, it represents something of a best case scenario.  Eq. \ref{eq:plane_slice_error} indicates that linearization error approaches zero as $s_1 \rightarrow s_2$.  This would not be the case, were the linear segments $L_1$ and $L_2$ to be separated by a non-zero angle in the local planar coordinate frame.  This scenario is illustrated by Fig. \ref{fig:sph_triangle_tangent_plane}.  Despite the fact that $(x_1,y_1)$ and $(x_2,y_2)$ appear to lie roughly the same distance from the origin, it is intuitively clear that the distance between the projected points ($\Delta L$) does not equal the true separation ($c$) between points $(lon_1,lat_1)$ and $(lon_2,lat_2)$.  Accordingly, it is important to account for the in-plane angular separation between points, $\gamma$, when characterizing linearization error.

We first require an expression for $\Delta L$ which accounts for a non-zero angular separation between co-planar points.  A reformulation of the Pythagorean Theorem gives Eq. \ref{eq:plane_dL}, which describes the linear separation between two co-planar points, as a function of their angular separation and distance from the origin.  For $\gamma=0$, $\Delta L = L_2 - L_1$, as was the case for two points on a great circle.  For $\gamma=\pi$, however, the two segments form a single straight line, and the projected separation between points is given by $\Delta L = L_1 + L_2$.

\begin{equation}
    \Delta L = \sqrt{L_1^2 + L_2^2 - 2 L_1 L_2 \cos{(\gamma)}}
    \label{eq:plane_dL}
\end{equation}

Next, the spherical law of cosines is used to determine the true separation between the points.  In Fig. \ref{fig:sph_triangle_tangent_plane}, this distance is represented by the arc length, $c$.

\begin{equation} 
    c = R \arccos{\left[ \cos{\left( \frac{s_1}{R} \right)}\cos{\left( \frac{s_2}{R} \right)} + 
    \sin{\left( \frac{s_1}{R} \right)} \sin{\left( \frac{s_2}{R} \right)} \cos{(\gamma)} \right]}
\end{equation}

It is convenient to express this distance in terms of the linear distance from the tangent plane origin.  Using a reformulation of Eq. \ref{eq:plane_proj}, the following expression is attained:

\begin{gather} 
    \Delta s = c = R \arccos{\left[ \frac{\sqrt{(R^2-L_1^2)(R^2-L_2^2)} + L_1 L_2 \cos{(\gamma)}}{R^2} \right]} \label{eq:plane_c} \\
    \epsilon = \Delta s - \Delta L \label{eq:plane_error}
\end{gather}

\begin{figure}[ht]
    \centering
    \includegraphics[width=0.5\textwidth]{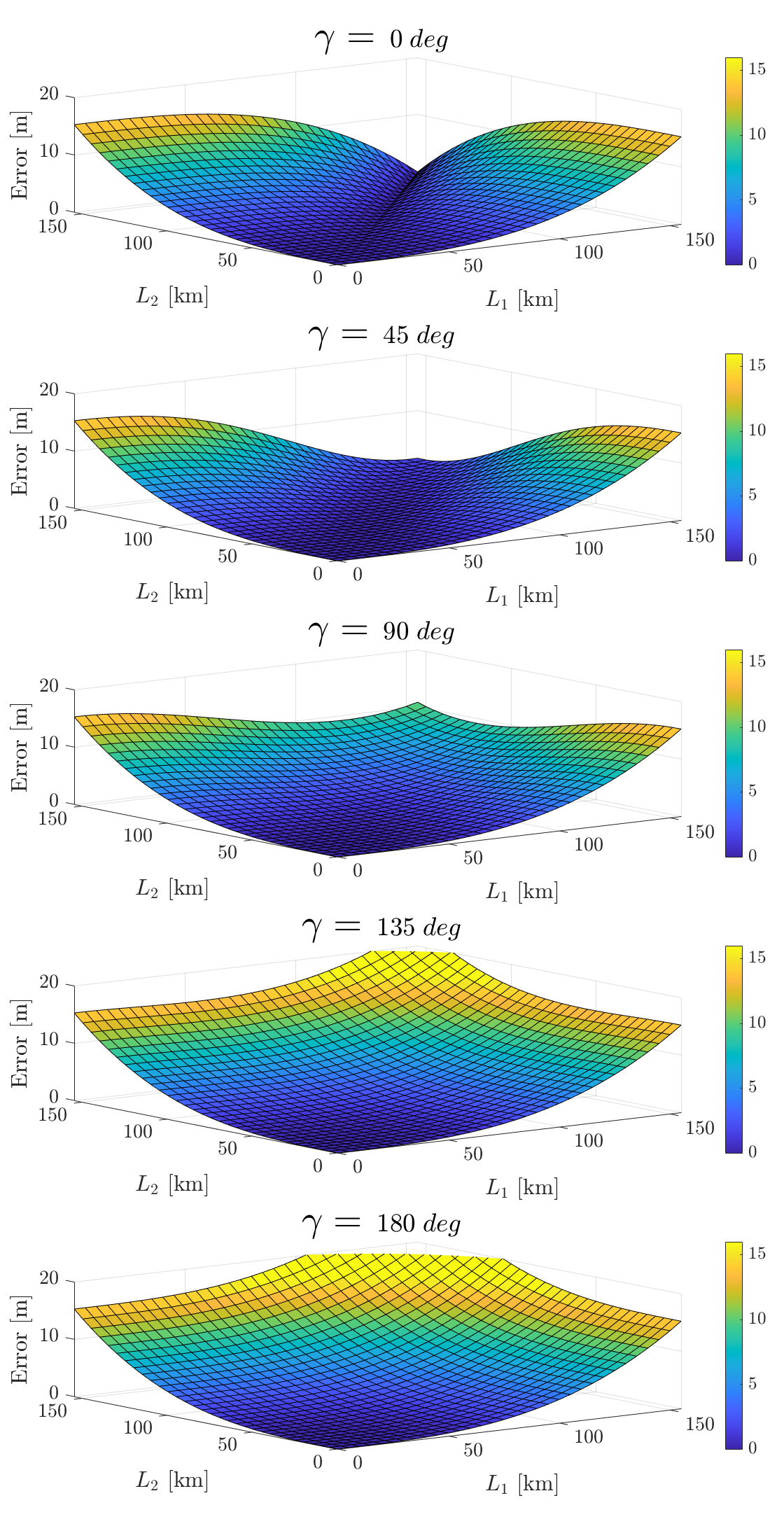}
    \caption{Separation error incurred by the projection of two geodetic positions onto a planar coordinate frame.  $L_1$ and $L_2$ represent the linear distance of each projected point from the tangent plane origin, while $\gamma$ describes the in-plane angular separation between the projected points.  The z-axis (and color axis) show the difference between the true distance between the two points (an arc on a sphere), and the projected point separation (a line segment on a plane).  This visualization of Eqs. \ref{eq:plane_dL}, \ref{eq:plane_c}, and \ref{eq:plane_error} underscores the influence of in-plane angular separation on range estimation error.}
    \label{fig:tangent_plane_error}
\end{figure}

Eqs. \ref{eq:plane_dL}, \ref{eq:plane_c}, and \ref{eq:plane_error} give a complete expression for linearization error, as a function of three variables: $L_1$, $L_2$, and $\gamma$.  The shape of the error function is strongly dependent on angular separation.  For small values of $\gamma$, and $L_1 \approx L_2$, linearization error remains small regardless of distance from the origin.  However, if $\gamma$ is large, the function becomes monotonic: any increase in $L_1$ or $L_2$ leads to an increase in error.  This behavior is shown in Fig. \ref{fig:tangent_plane_error}.  In general, an increase in angular separation leads to an increase in linearization error, for a given combination of $L_1$ and $L_2$.  Thus, the case of $\gamma = \pi$ gives a conservative estimate of separation error due to linearization, and should be used to determine when it is necessary to define a new planar coordinate system.

Eqs. \ref{eq:plane_dL}, \ref{eq:plane_c}, and \ref{eq:plane_error} indicate that so long as the projected distance from each point to the origin remains less than 100 kilometers, the incurred projection error will be less than 8.3 meters; in order to limit error to 1 meter, each contact must be less than 50km from the origin.  It is not difficult to see how this could be undesirable, if not problematic, for long-distance voyages.  For example, if an ASV operator wanted to ensure that linearization error remained less than 1 meter throughout a trans-Atlantic voyage of 2,992 nautical miles, the ASV would need to be programmed to redefine its local planar coordinate frame no fewer than 110 times.  This is not so much a computational issue as a logistical one.  This task becomes even more challenging if a group of vessels are travelling together in the same general direction as the ASV.  If these vessels frequently pass in and out of the valid region of the local coordinate system, they will repeatedly trigger automated redefinitions of the local frame.  This will both obfuscate the operating region, and incur needless computational overhead.  Further, what if there does not exist a local frame capable of keeping all tracked vessels (and ownship) within the valid region of that frame?  Continuous target tracking over multiple discontinuous local coordinate frames is a non-trivial problem, and gives rise to a multitude of additional software requirements.  These complications can be circumvented by tracking targets directly in geodetic coordinates.

\section{Uniform Distribution of Points on a Sphere}
\label{appendix:uniform_sphere_points}
A uniform distribution of latitude and longitude does not yield a uniform distribution of points on a spherical shell.  Test points selected in this manner will lead to over-sampling at higher latitudes, and under-sampling near the equator, due to the shrinking circumferential length of longitudinal rings with increasing latitude.  The following algorithm rectifies this issue, and yields a set of evenly-distributed random geodetic points on the surface of a sphere.

\begin{align}
    npts &= 1e^{+07} \notag\\
    \boldsymbol{u} &= rand(npts) \notag\\
    \boldsymbol{v} &= rand(npts) \notag\\
    \boldsymbol{lon} &= \left(\frac{180}{\pi} \right) \left(2 \pi \boldsymbol{u}  \right)\\
    \boldsymbol{lat} &= \left(\frac{180}{\pi} \right) \left(\arccos{(2\boldsymbol{v}-1)} - \frac{\pi}{2}  \right)
\end{align}

\section{A Remark on AIS Spoofing}
A common critique of AIS is its vulnerability to \textit{spoofing} attacks.  Spoofing is a term used to describe the intentional, malicious transmission of false vessel data over AIS frequencies.  This is possible due to the fact that AIS is not a secure protocol.  The most common motivations for AIS spoofing include masking illegal fishing operations, and provoking adversarial naval forces.  This shortcoming has been highly publicized in mainstream media outlets \citep{harris2019}, and covered extensively in the technical literature \citep{katsilieris2016, kontopoulos2018}; however, such incidents are relatively infrequent in practice, and of interest primarily to coastal defense, immigration, and regulatory agencies.  Under most circumstances, AIS provides a rich stream of valid information, at no cost to the user.  If leveraged properly, AIS can drastically enhance the collision avoidance capabilities of an ASV.

\newpage
\section{AIS Message Format}
\label{appendix:AIS}
Decoding an AIS message is not entirely trivial.  After verifying the integrity of the raw NMEA payload using the appended checksum, one must perform a number of steps.  First, fragmented AIS messages must be combined; such multi-part messages are common for lengthy \textit{Type 5} messages carrying static vessel information.  Once the complete payload has been extracted, each character therein must be converted to its corresponding ASCII value; then, each ASCII value is converted to a decimal value; finally, each decimal value is converted to a 6-bit binary value.  Once the AIS message contains only binary values, the bits must be partitioned according to the AIS message type.

\begin{table}[ht]
\caption{Dynamic AIS Message Binary Bit Map}
\label{tab:dynamic_AIS_bits}
\centering
\begin{tabular}{lll}
\toprule
        & \textbf{Type 1, 2, 3}  & \textbf{Type 18} \\
        & [Class A]  & [Class B] \\
\midrule
\textbf{Repeat Count}       & 6-7       & 6-7    \\
\textbf{MMSI Vessel ID}     & 8-37      & 8-37   \\
\textbf{Navigation Status}  & 38-41     & --     \\
\textbf{Turn Rate}          & 42-49     & --     \\
\textbf{Speed Over Ground}  & 50-59     & 46-55  \\
\textbf{GNSS Accuracy}      & 60        & 56     \\
\textbf{Longitude}          & 61-88     & 57-84  \\
\textbf{Latitude}           & 89-115    & 85-111 \\
\textbf{Course Over Ground} & 116-127   & 112-123\\
\textbf{Heading}            & 128-136   & 124-132\\
\textbf{Time Stamp [sec]}   & 137-142   & 133-138\\
\textbf{Maneuver Code}      & 143-144   & --     \\
\bottomrule
\end{tabular}
\end{table}

There are two primary AIS message types: static and dynamic.  While the former contain valuable information like the vessel name and dimensions, the latter are of primary interest for vessel tracking.  Dynamic messages are published at various frequencies, depending on the size of the vessel, its speed, and whether or not it is maneuvering.  For the most part, large vessels with Class A AIS transponders publish their kinematic states every 2 - 10 seconds, while smaller vessels with Class B transponders transmit every 30 seconds.

\begin{table}[ht]
\caption{Class A AIS Reporting Intervals}
\label{tab:AIS_freq}
\centering
\begin{tabular}{lllll}
\toprule
       & \textbf{Class} & \textbf{Speed [kts]} & \textbf{Constant Course} & \textbf{Changing Course}  \\
\midrule
\textbf{At anchor or moored}   & A & $<$ 3  & 3 min  & 3 min    \\
\textbf{Under way}             & A & 0-14   & 10 sec  & 3 1/3 sec \\
\textbf{Under way}             & A & 14-23  & 6 sec   & 2 sec \\
\textbf{Under way}             & A  & $>$ 23  & 2 sec  & 2 sec \\
\textbf{Under way}             & B  & $<$ 2   & 3 min  & 3 min \\
\textbf{Under way}             & B  & 2-14    & 30 sec  & 30 sec \\
\textbf{Under way}             & B  & 14-23    & 15 sec  & 15 sec \\
\textbf{Under way}             & B  & $>$ 23    & 5 sec  & 5 sec \\
\bottomrule
\end{tabular}
\end{table}

After partitioning the bits according to Table \ref{tab:dynamic_AIS_bits}, the binary values must be converted back to decimal values.  Each variable then requires additional processing (sometimes referred to as ``unpacking''), as described in Table \ref{tab:AIS_postprocessing}.

\begin{table}
\caption{AIS Field Post-Processing Requirements}
\label{tab:AIS_postprocessing}
\centering
\begin{tabular}{llll@{} }
\toprule
        & \textbf{NaN Value}  & \textbf{Unpacking Operation}    & \textbf{Units}\\
\midrule
\textbf{Longitude}          & 181           & $6e^{-5}x$            & deg \\
\textbf{Latitude}           & 91            & $6e^{-5}x$            & deg \\
\textbf{Speed Over Ground}  & 1023          & $\frac{0.51444x}{10}$ & m/s \\
\textbf{Course Over Ground} & [3600:4095)   & $\frac{x}{10}$        & deg \\
\bottomrule
\end{tabular}
\end{table}

Decoding the vessel name -- contained in each static AIS message (Type 5) -- is interesting, and facilitates validation of AIS data with external sources.  Should the user be interested in decoding this field, they must first partition the bits according to Table \ref{tab:static_AIS_bits}, and then apply the following procedure to the \textit{Vessel Name} bits:
\begin{enumerate}
    \item Divide the payload (bits 112-231) into 6-bit chunks.
    \item Convert each 6-bit byte into its corresponding decimal value.
    \item Remove trailing zeros.
    \item Convert decimal values to ASCII characters.
    \item Trim white space.
\end{enumerate}

\begin{table}
\caption{Static AIS Message Binary Bit Map}
\label{tab:static_AIS_bits}
\centering
\begin{tabular}{ll}
\toprule
        & \textbf{Type 5}  \\
        & [Class A]                    \\
\midrule
\textbf{Repeat Count}       & 6-7       \\
\textbf{MMSI Vessel ID}     & 8-37      \\
\textbf{IMO Vessel ID}      & 40-69     \\
\textbf{Vessel Name}        & 112-231   \\
\textbf{Vessel Type Code}   & 232-239   \\
\textbf{Distance To Bow}    & 240-248   \\
\textbf{Distance To Stern}  & 249-257   \\
\textbf{Distance To Port}   & 258-263   \\
\textbf{Distance To Starboard}& 264-269 \\
\textbf{Vessel Draught}     & 294-301   \\
\textbf{GNSS Fix Type}      & 270-273   \\
\bottomrule
\end{tabular}
\end{table}

\section{Nomenclature}
\begin{table}[ht]
\caption{Abbreviations}
\label{tab:abbreviations}
\centering
\begin{tabular}{ll}
\toprule
\textbf{ASV}        & Autonomous Surface Vehicle \\
\textbf{COLREGS}    & Convention on the International Regulations for Preventing Collisions at Sea \\
\textbf{RADAR}      & RAdio Detection And Ranging \\
\textbf{LIDAR}      & Light Detection and Ranging \\
\textbf{VHF}        & Very High Frequency\\
\textbf{IMO}        & International Maritime Organization\\
\textbf{SOLAS}      & Safety of Life at Sea\\
\textbf{V2V}        & Vehicle to Vehicle\\
\textbf{SOG}        & Speed Over Ground \\
\textbf{COG}        & Course Over Ground \\
\textbf{MMSI}       & Maritime Mobile Service Identity\\
\textbf{k-NN}       & k-Nearest Neighbors \\
\textbf{LKF}        & Linear Kalman Filter \\
\textbf{EKF}        & extended Kalman filter \\
\textbf{XKF}        & eXogenous Kalman Filter \\
\textbf{UKF}        & unscented Kalman filter \\
\textbf{ARE}        & Algebraic Riccati Equation \\
\textbf{CV}         & Constant Velocity (Motion Model) \\
\textbf{CA}         & Constant Acceleration (Motion Model) \\
\textbf{CT}         & Constant Turn (Motion Model) \\
\textbf{ECEF}       & Earth-Centered Earth-Fixed (Coordinates) \\
\textbf{NED}        & North-East Down (Coordinates) \\
\textbf{RMSE}       & Root Mean Squared Error \\
\textbf{UART}       & Universal Asynchronous Receiver-Transmitter \\
\textbf{GPIO}       & General Purpose Input-Output \\
\bottomrule
\end{tabular}
\end{table}

\bibliographystyle{unsrtnat}

\clearpage
\bibliography{main}






\end{document}